\pdfoutput=1

\documentclass[11pt]{article}

\usepackage[preprint]{acl}

\usepackage{times}
\usepackage{latexsym}

\usepackage[T1]{fontenc}

\usepackage[utf8]{inputenc}

\usepackage{microtype}

\usepackage{inconsolata}

\usepackage{inconsolata}
\usepackage{amsmath,graphicx}
\usepackage{tabularx}
\usepackage{amssymb}
\usepackage{amsthm}
\usepackage{cleveref}
\usepackage{pifont}
\usepackage{cancel}
\usepackage{lipsum}
\usepackage{fvextra}
\usepackage{booktabs}
\usepackage{algorithm}
\usepackage{algorithmic}
\usepackage{multirow}
\usepackage{enumitem}
\usepackage{ragged2e}
\usepackage{subcaption}
\usepackage{graphicx}
\usepackage{booktabs}
\usepackage{multirow}
\usepackage[most]{tcolorbox}
\usepackage{titling}
\usepackage{pifont}
\usepackage{float}
\usepackage{xcolor}

\title{Discrete Minds in a Continuous World: \\ Do Language Models Know Time Passes?}

\author{Minghan Wang, Ye Bai, Thuy-Trang Vu, \textbf{Ehsan Shareghi}, \textbf{Gholamreza Haffari} \\
  Department of Data Science \& AI, Monash University \\
  \texttt{\{firstname.lastname\}}@monash.edu
}

\begin{document}
\maketitle
\begin{abstract}
While Large Language Models (LLMs) excel at temporal reasoning tasks like event ordering and duration estimation, their ability to perceive the actual passage of time remains unexplored. We investigate whether LLMs perceive the passage of time and adapt their decision-making accordingly through three complementary experiments. First, we introduce the Token-Time Hypothesis, positing that LLMs can map discrete token counts to continuous wall-clock time, and validate this through a dialogue duration judgment task. Second, we demonstrate that LLMs could use this awareness to adapt their response length while maintaining accuracy when users express urgency in question answering tasks. Finally, we develop BombRush, an interactive navigation challenge that examines how LLMs modify behavior under progressive time pressure in dynamic environments. Our findings indicate that LLMs possess certain awareness of time passage, enabling them to bridge discrete linguistic tokens and continuous physical time, though this capability varies with model size and reasoning abilities. This work establishes a theoretical foundation for enhancing temporal awareness in LLMs for time-sensitive applications.\footnote{Code and data will be released upon acceptance.}
\end{abstract}

\section{Introduction}

\begin{figure}[t]
    \centering
    \includegraphics[width=\columnwidth]{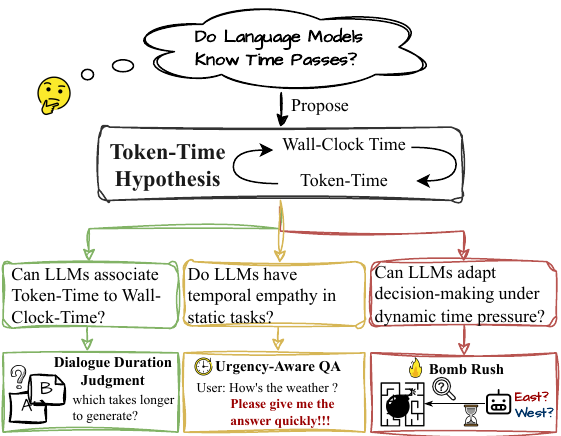}
    \caption{Overview of our work. We propose the Token-Time Hypothesis connecting discrete tokens to continuous time, then validate it through three experiments examining time-mapping capabilities, temporal empathy in static tasks, and adaptive decision-making under dynamic tasks.}
    \label{fig:overview}
    \vspace{-1em}
\end{figure}

Current LLMs are predominantly employed for \textbf{time-insensitive} tasks such as question answering (QA)~\citep{hendrycks2021measuringmassivemultitasklanguage,rein2023gpqagraduatelevelgoogleproofqa}, summarization~\citep{zhang2023benchmarkinglargelanguagemodels}, and translation~\citep{xu2024contrastivepreferenceoptimizationpushing}. Although both user input and LLM generation inherently consume time, real-world temporal factors typically do not significantly influence these tasks' outcomes. For instance, neither the duration taken by a user to formulate a question nor the LLM's response latency substantially alters the correctness of the final output.

However, numerous \textbf{time-sensitive} tasks exist in real-world scenarios, including simultaneous translation~\citep{wang2024simultaneousmachinetranslationlarge,wang2024conversationalsimulmtefficientsimultaneous}, autonomous driving~\citep{yang2024llm4drivesurveylargelanguage}, real-time dialogue~\citep{défossez2024moshispeechtextfoundationmodel}, and robotic control~\citep{driess2023palmeembodiedmultimodallanguage}. In these contexts, environmental conditions continually evolve over time, resulting in input data variability and task dynamics heavily dependent on temporal progression. Moreover, the time taken by a model to generate responses can critically impact task outcomes. Hence, temporal progression becomes a decisive factor in these scenarios. This raises a fundamental question: \textit{Are LLMs capable of perceiving and interpreting the passage of time? Specifically, can LLMs recognize their temporal context and adapt their behavior accordingly?}

To date, extensive research has focused on LLMs' reasoning capabilities in time-related tasks, such as event ordering~\citep{zhou2021temporalreasoningimplicitevents,tan2023benchmarkingimprovingtemporalreasoning,ding2025do}, temporal expression parsing~\citep{chen2021datasetansweringtimesensitivequestions,zhang2021situatedqaincorporatingextralinguisticcontexts,zhou-etal-2019-going}, and duration computation~\citep{wang2024trambenchmarkingtemporalreasoning,Jia_2018,shang2022improvingtimesensitivityquestion, mavromatis2021tempoqrtemporalquestionreasoning}. These studies evaluate models’ ability to perform temporal reasoning \textbf{but do not address whether models possess any awareness of the passage of time itself within our physical world}.

In this paper, we bridge this gap by investigating whether and how LLMs perceive the passage of time and adapt their behavior accordingly. We first analyze the mechanisms through which LLMs might understand time and propose the \textbf{Token-Time Hypothesis}: LLMs can establish connections between text length and real-world temporal progression. Through three complementary experiments,  we systematically explore the nature and boundaries of this capability.

Our investigation begins with a Dialogue Duration Judgment task that validates the Token-Time Hypothesis by examining whether LLMs can accurately determine which conversation took longer based on different temporal cues. We find that models can indeed associate token count with elapsed time, though their ability to reconcile conflicting cues depends significantly on their reasoning capabilities. Building on these insights, we then explore how this temporal awareness manifests in practical scenarios through an Urgency-Aware QA experiment, which reveals that models strategically reduce response length when users express time constraints while preserving answer quality, demonstrating a form of "temporal empathy." Finally, our BombRush experiment extends this investigation to dynamic, interactive environments, showing how models adjust their reasoning depth and decision-making strategies under progressively increasing time pressure.

This work makes several key contributions to our understanding of language model capabilities: \textbf{(1)} We present the first comprehensive study investigating LLMs' perception of temporal progression. \textbf{(2)} We establish the Token-Time Hypothesis as a theoretical framework explaining how discrete linguistic tokens might be mapped to continuous temporal experience. \textbf{(3)} We design three novel experimental paradigms to evaluate how time passage perception influences LLM behavior across diverse scenarios. \textbf{(4)} We provide empirical evidence that LLMs possess certain temporal passage awareness, though varying across model scales and genres. These findings have important implications for deploying LLMs in time-sensitive applications and provide a foundation for future research on enhancing temporal awareness in language models.




\section{LLM's Sense of Time}
\begin{figure*}[t]
    \centering
    \includegraphics[width=\textwidth]{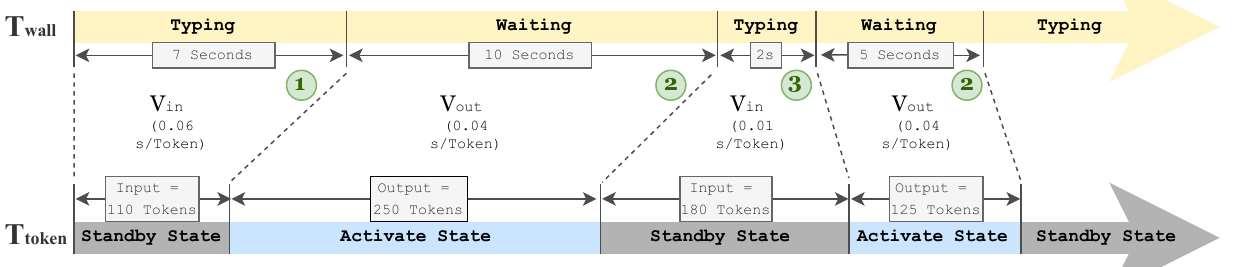}
    \caption{Illustration of the relationship between Token Time and Wall-Clock-Time across LLM's operational cycles. This figure compares the Wall-Clock timeline ($T_{wall}$) and the model’s Token timeline ($T_{token}$). The LLM alternates between a \textbf{Standby State} (waiting for user input) and an \textbf{Activate State} (generating responses). While the LLM can observe the number of input tokens, it cannot infer how long the user spent composing them, as input speed is unobservable and variable. For example, in Case \textcolor{green}{\scalebox{1.1}{\ding{172}}}, the user typed slowly while thinking, yielding few tokens over a long period. In Case \textcolor{green}{\scalebox{1.1}{\ding{174}}}, the user quickly pasted a long passage, resulting in many tokens in a short time. In contrast, during the generation phase, the LLM's output token count provides a measurable passage of time, assuming an ideally fixed generation speed ($V_{out}$), as shown in Case \textcolor{green}{\scalebox{1.1}{\ding{173}}}. \textbf{This enables an estimation of elapsed Wall-Clock Time based on the number of output tokens.}}
    \label{fig:rethinking_timeline}
\vspace{-1em}
\end{figure*}

Time perception (chronoception) in psychology refers to the subjective experience of time: how individuals sense durations and intervals between events, distinct from objective chronological time~\citep{10.3389/fnhum.2018.00074,EAGLEMAN2008131,citeulike:4758119}.

Building on this concept, we investigate whether similar mechanisms exist in AI systems. Autoregressive decoder-only language models process text by decomposing sequences into conditional probability distributions that preserve temporal ordering~\citep{brown2020languagemodelsfewshotlearners}. This creates a natural alignment between token positions and chronological progression: earlier tokens precede later ones, mirroring how events unfold in human experience. Moreover, token quantity roughly corresponds to communication time in human interaction. While humans intuitively incorporate temporal dimensions in communication, a fundamental question emerges: do LLMs possess an analogous sense of time that influences their comprehension and generation?

\subsection{Time Perception Mechanisms in LLMs}

To systematically investigate potential temporal awareness in language models, we focus on the most representative scenario, i.e. multi-turn dialogue, where time perception becomes particularly relevant. Conversations provide an ideal context due to their inherent chronological structure, participants exchange messages sequentially, creating a natural temporal flow that allows us to examine how LLMs perceive and process time-related information.

Within this interactive framework, an LLM's operational cycle can be abstracted into two states: \textbf{Standby State} (awaiting input without computation) and \textbf{Active State} (processing input through generation completion). During standby, the model exists in temporal isolation, with time perception potentially manifesting only in two scenarios:
\begin{enumerate}[noitemsep,topsep=2pt,parsep=2pt,partopsep=2pt]
    \item \textbf{During Encoding of User Input}: When activated by new input, the model encodes this text, potentially extracting temporal information from either the number of input tokens or explicit time markers within the input. This mechanism allows the model to infer time that elapsed during its standby period.
    \item \textbf{During Autoregressive Generation}: While generating text, the model cannot incorporate new external input, but theoretically could estimate passing wall-clock time if it possesses awareness of its own generation speed and the relationship between the number of generated tokens and time elapsed.
\end{enumerate}

\subsection{The Token-Time Hypothesis}

These observations lead us to formulate the \textbf{Token-Time Hypothesis}, which proposes that: \textit{LLMs treat tokens as discrete temporal units, inferring the passage of real-world time from the length and sequencing of textual events within the token space}. This hypothesis establishes two distinct temporal measurement systems: \textbf{Token-Time}, the discrete, abstract temporal metric based on token counts, and \textbf{Wall-Clock-Time}, the continuous, physical temporal metric in the real world. Under this framework, studying an LLM's temporal awareness fundamentally involves understanding how the model perceives and utilizes the mapping relationship between these two temporal domains. \textbf{This mapping capability, if present, would be crucial for LLMs to function effectively in time-sensitive applications where understanding temporal progression impacts task performance.} (More discussions can be found in~\Cref{sec:app_tth_connection}.)

\subsection{Mapping Between Token-Time and Wall-Clock-Time}
To formalize the relationship between these temporal domains, we define a mathematical framework that quantifies this critical mapping:

\begin{align}
T_{wall}^{m} = T_{tok}^{m} \times V_{m}, \quad m \in \{in, out\}
\end{align}
where $V_{m}$ represents the conversion rate between domains for mode $m$ (input or output), measured in seconds per token (s/token). Specifically, the input conversion rate $V_{in}$ is calculated as the ratio of Wall-Clock time spent on input preparation $T_{wall}^{in}$ to the total token count of that input $T_{tok}^{in}$. Similarly, the output conversion rate $V_{out}$ represents the ratio of actual generation time $T_{wall}^{out}$ to the number of tokens produced $T_{tok}^{out}$.

In ideal conditions, when both $V_{in}$ and $V_{out}$ remain constant, the relationship $T_{wall} \propto T_{tok}$ would hold. However, real-world scenarios demand a more nuanced consideration, particularly for $V_{in}$. In multi-turn dialogues, various external factors (as illustrated in \Cref{fig:rethinking_timeline}) cause $V_{in}$ to fluctuate significantly, preventing a consistent $T_{wall}^{in} \propto T_{tok}^{in}$ relationship. Conversely, $V_{out}$ remains relatively stable (assuming negligible computational overhead from increasing generation length), maintaining the proportionality $T_{wall}^{out} \propto T_{tok}^{out}$. To ensure experimental rigor, we control for input variability and focus primarily on LLMs' understanding and utilization of time passage during the Activate State (output) phase.

Specifically, to address our research questions, we conducted three experiments. The first experiment validates whether LLMs can comprehend the $T_{wall}^{out} \propto T_{tok}^{out}$ mapping relationship. In the subsequent two experiments, we investigate whether LLMs with time passage perception capabilities demonstrate behavior that strategically controls $T_{tok}^{out}$ to influence $T_{wall}^{out}$, thereby meeting the temporal requirements of particular tasks.

\section{Dialogue Duration Judgment}
\label{sec:pre_task}
\begin{figure}[t]
    \centering
    \includegraphics[width=\columnwidth]{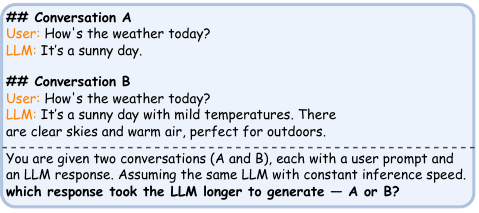}
    \caption{Prompt example for the Dialogue Duration Judgment task. LLMs are given two user–LLM conversations and asked to judge which response took the LLM longer to generate.
    }
    \label{fig:ddj_example}
\vspace{-1em}
\end{figure}

\begin{table*}[t]
\centering
\resizebox{\textwidth}{!}{%
\begin{tabular}{@{}lll@{}}
\toprule
\textbf{Setting} & \textbf{Key Characteristics} & \textbf{Purpose} \\ \midrule
\textbf{S1} & No explicit temporal cues provided & Test inherent ability to infer temporal duration from $T_{tok}$ \\
\textbf{S1-Hint} & Textual hint: ``Generation time is proportional to number of tokens'' & Evaluate if the model understands the hint: $T_{tok} \propto T_{wall}$ \\
\textbf{S1-Count} & Direct token counts provided for both dialogues & Test ability to judge duration from explicit $T_{tok}$ information \\ \midrule
\textbf{S2} & Timestamps provided for input and response; consistent with text length & Test prioritization of explicit $T_{wall}$ cues over implicit $T_{tok}$ \\
\textbf{S2-M} & Misleading timestamps: longer responses shown with shorter durations & Evaluate ability to prioritize $T_{wall}$ over contradictory $T_{tok}$ cues \\
\textbf{S2-M+} & Both misleading timestamps and accurate token counts provided & \begin{tabular}[c]{@{}l@{}}Determine which temporal domain ($T_{tok}$ vs $T_{wall}$) \\ models prioritize when directly conflicting\end{tabular} \\ \bottomrule
\end{tabular}%
}
\caption{Experimental settings for the Dialogue Duration Judgment task. Settings are divided into Token-Time cues (S1, S1-Hint, S1-Count) and Wall-Clock-Time cues (S2, S2-M, S2-M+).}
\label{tab:ddj_exp_settings}
\end{table*}
Our first experiment examines whether LLMs can associate text length (Token-Time) with temporal duration (Wall-Clock-Time), and how this capability functions under various confounding factors. To systematically evaluate this, we designed a \textbf{dialogue duration judgment} task using paired samples from the  chatbot\_arena\_conversations dataset\footnote{lmsys/chatbot\_arena\_conversations}~\citep{zheng2023judgingllmasajudgemtbenchchatbot}. Each sample contains a user prompt with two responses (A or B) from different LLMs of substantially different lengths. We selected 300 samples where both responses were generated by high-quality models such as GPT-4 or Claude, ensuring response quality would not negatively affect LLMs' decision-making. The tested LLM must identify which dialogue is likely to require more time to complete, and provide justification for its judgment (See \Cref{fig:ddj_example} for an example).

\subsection{Experimental Setup}
We systematically designed six controlled settings by introducing different temporal cues that either facilitate or impede the LLM's judgment. These settings are organized into two categories: \textbf{Token-Time cues} and \textbf{Wall-Clock-Time cues}, each probing whether the model could correctly leverage the corresponding temporal domain to make accurate predictions. The detailed task prompts and demonstration of settings are illustrated in~\Cref{ddj_appendix_setting},~\ref{ddj_appendix_prompts} and~\Cref{tab:ddj_exp_settings}.

\vspace{-1mm}
\paragraph{Setting 1: Token-time Cues}
For this group of settings, we progressively introduced more explicit cues to help the candidate LLM establish the connection between Token-Time and Wall-Clock-Time. To eliminate potential confounding factors, we explicitly stated in the prompt that all responses were generated by an identical model with constant inference speed.

\vspace{-1mm}
\paragraph{Setting 2: Wall-Clock-Time Cues}
In this group, we introduced temporal logs containing explicit Wall-Clock-Time timestamps for both dialogues to test whether the LLM could effectively utilize these direct temporal signals. We also manipulated these timestamps in certain settings to create conflicting cues, allowing us to observe how LLMs prioritize different temporal indicators. These manipulations simulate realistic scenarios where varying computational conditions might create discrepancies between text length and generation time.

This experimental design probes how LLMs negotiate between different temporal metrics and whether they possess coherent mechanisms for temporal perception.

\vspace{-1mm}
\paragraph{Models and Evaluation}
\label{sec:ddj_model_eval}
We tested six models divided into three categories: Small LMs (7-8B parameters): Qwen-2.5-7B\footnote{Qwen/Qwen2.5-7B-Instruct}~\citep{qwen2.5} and Llama-3.1-8B\footnote{meta-llama/Llama-3.1-8B-Instruct}~\citep{grattafiori2024llama3herdmodels}; Large LMs (70B+ parameters): Llama-3.3-70B\footnote{meta-llama/Llama-3.3-70B-Instruct} and Qwen-2.5-72B\footnote{Qwen/Qwen2.5-72B-Instruct}; and Large Reasoning Models (LRMs with internalized inference time scaling capabilities): DeepSeek-R1-Distill-Llama-70B\footnote{deepseek-ai/DeepSeek-R1-Distill-Llama-70B}~\citep{deepseekai2025deepseekr1incentivizingreasoningcapability} and QWQ-32B\footnote{Qwen/QwQ-32B}~\citep{qwq32b}. For each model under each setting, we conducted 5 replications and reported the average accuracy as our primary evaluation metric. To ensure statistical rigor, we further conducted 20 replications for each model-setting combination, with detailed significance testing in~\Cref{ddj_appendix_ttest}.

\subsection{Results}
\begin{table}[t]
\centering
\resizebox{\columnwidth}{!}{%
\begin{tabular}{ll|rrr|rrr}
\hline
 & \textbf{Settings} & \multicolumn{1}{c}{\textbf{S1}} & \multicolumn{1}{c}{\textbf{S1-Hint}} & \multicolumn{1}{c|}{\textbf{S1-Count}} & \multicolumn{1}{c}{\textbf{S2}} & \multicolumn{1}{c}{\textbf{S2-M}} & \multicolumn{1}{c}{\textbf{S2-M+}} \\ \hline
\multirow{2}{*}{\textbf{SLM}} & \textbf{Llama-8B} & 79.7 & 83.0 & \textbf{99.4} & 76.8 & 47.7 & 16.3 \\
 & \textbf{Qwen-7B} & 66.8 & 71.2 & \textbf{92.7} & 80.9 & 26.8 & 12.9 \\
\multirow{2}{*}{\textbf{LLM}} & \textbf{Llama-70B} & 91.3 & 93.2 & \textbf{99.9} & 93.5 & 74.0 & 40.1 \\
 & \textbf{Qwen-72B} & 83.8 & 84.5 & \textbf{99.9} & 96.6 & 82.5 & 48.8 \\
\multirow{2}{*}{\textbf{LRM}} & \textbf{DLlama-70B} & 87.9 & 92.3 & \textbf{100.0} & 99.2 & 98.0 & 94.3 \\
 & \textbf{QwQ-32B} & 91.5 & 94.8 & \textbf{100.0} & 99.3 & 99.2 & 99.1 \\ \hline
\end{tabular}%
}
\caption{Accuracy (\%) of different model categories on the Dialogue Duration Judgment task. Settings include Token-Time cues (S1: baseline, S1-Hint: with hint, S1-Count: with token counts) and Wall-Clock-Time cues (S2: consistent timestamps, S2-M: misleading timestamps, S2-M+: misleading timestamps with token counts).}
\label{tab:pre_task_result}
\end{table}

\begin{figure}[t]
    \centering
    \includegraphics[width=\columnwidth]{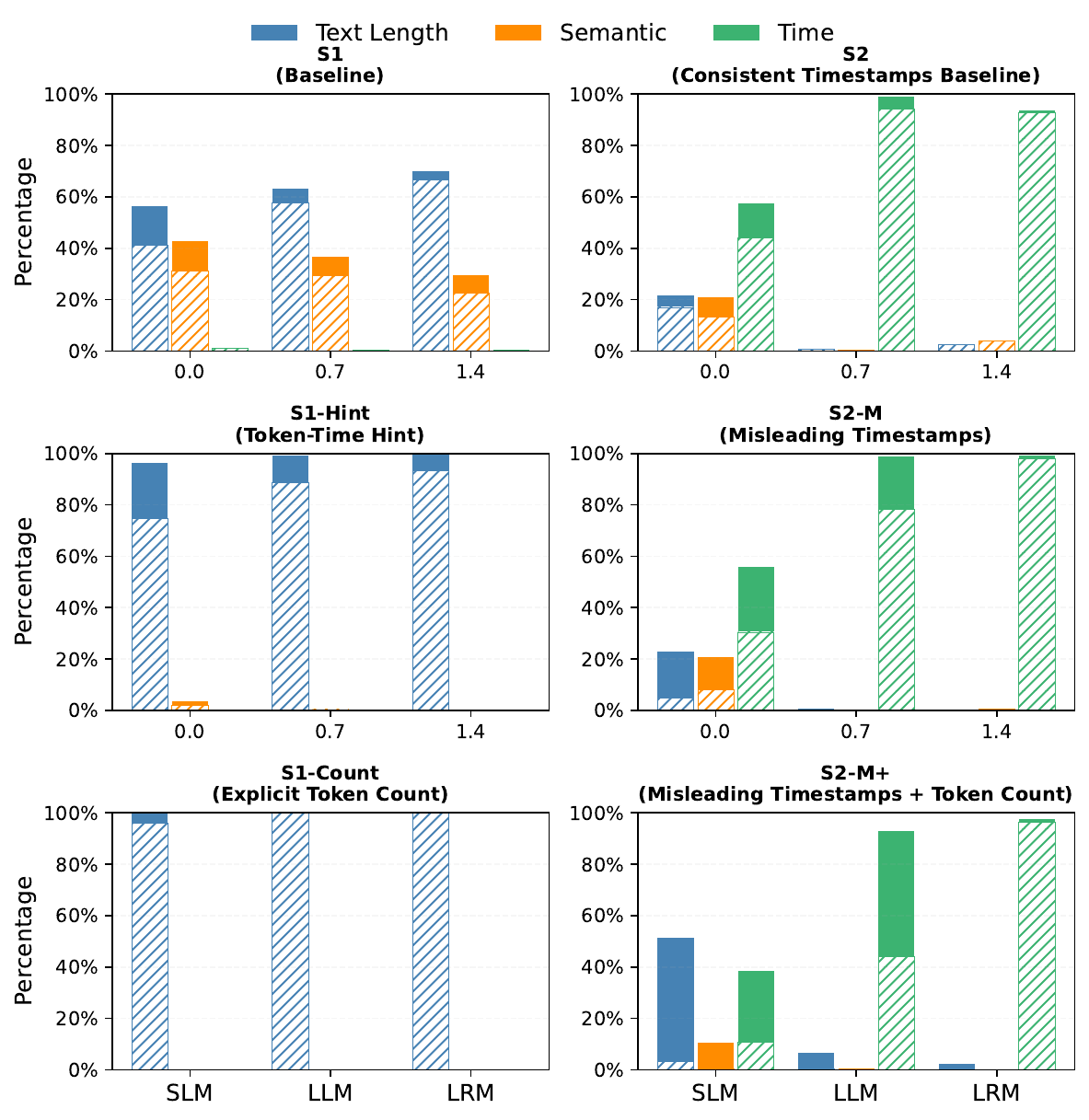}
    \caption{Attribution distribution and accuracy across model classes and settings. Each bar shows the proportion of attribution used, hatched segments indicate the percentage of correct answers within that attribution. (See \Cref{ddj_appendix_attri_analysis} for detailed numerical results)
    }
    \vspace{-1em}
    \label{fig:attribute_analysis}
\end{figure}

\Cref{tab:pre_task_result} presents the accuracy of all models across different experimental settings. We analyze performance by model category and within settings, followed by attribution analysis to understand underlying reasoning mechanisms. 

\vspace{-1mm}
\paragraph{Performance Across Model Categories}
We observe a clear progression in accuracy from SLMs to LLMs to LRMs, aligning with expectations that larger models and those with enhanced reasoning capabilities demonstrate superior overall performance, potentially including the ability to perceive the passage of time that we are investigating.

\vspace{-1mm}
\paragraph{Token-Time Settings (S1, S1-Hint, S1-Count)}
In the baseline condition (S1), all models achieve reasonably high accuracy without explicit cues, indicating an inherent capability to associate text length with temporal duration. As Token-time hints become more explicit, accuracy increases across all model categories, reaching near-perfect scores in S1-Count where token counts are explicitly provided.

\vspace{-1mm}
\paragraph{Wall-Clock-Time Settings (S2, S2-M, S2-M+)}
Most models achieve higher accuracy with explicit timestamps (S2) compared to the Token-time baseline (S1), indicating that direct temporal cues enhance duration judgment beyond default token-based heuristics.

The most revealing results emerge with misleading cues (S2-M and S2-M+). SLMs and LLMs show dramatic performance degradation when token counts and timestamps provide contradictory signals. In stark contrast, LRMs maintain remarkably high accuracy even under these challenging conditions, with QwQ-32B achieving 99.1\% accuracy in the most conflicting setting.

\subsubsection{Attribution Analysis}
To understand the underlying mechanisms of temporal judgment, we classified models' justifications into three categories: "text length," "semantic," and "time" using a Llama-3.3-70B classifier (see~\Cref{fig:attribute_prompt} for the prompt). \Cref{fig:attribute_analysis} shows attribution distributions and corresponding accuracy across settings and model categories. 

In the baseline setting (S1), all models primarily use text length for attribution, though with considerable reliance on semantic cues as well. Judgments based on text length achieve substantially higher accuracy than those based on semantic features. As Token-time hints become more explicit, models increasingly favor text length attributions over semantic cues. For Wall-Clock-Time settings, LLMs show greater aptitude than SLMs at leveraging timestamp information in S2. Interestingly, LRMs occasionally incorporate multiple attribution types, but mostly rely on timestamp cues.
Under misleading settings (S2-M), SLMs and LLMs maintain similar attribution patterns as in S2, but with significantly reduced accuracy. In the most challenging setting (S2-M+), both SLMs and LLMs are substantially misled, increasingly relying on text length despite contradictory timestamp information. Even when LLMs continue using timestamp cues, their accuracy declines dramatically, indicating confusion when faced with conflicting temporal signals. LRMs, however, remain largely resistant to these misleading cues. We speculated more about this phenomenon's underlying mechanism in~\Cref{sec:app_how_time_is_processed}.

\begin{tcolorbox}[colback=white,colframe=black!75!white,title=Key Takeaway]
LLMs can perceive the passage of time by connecting Token-Time with Wall-Clock-Time, although this intriguing ability varies with their underlying reasoning capabilities.
\end{tcolorbox}
\section{\includegraphics[height=1.5em]{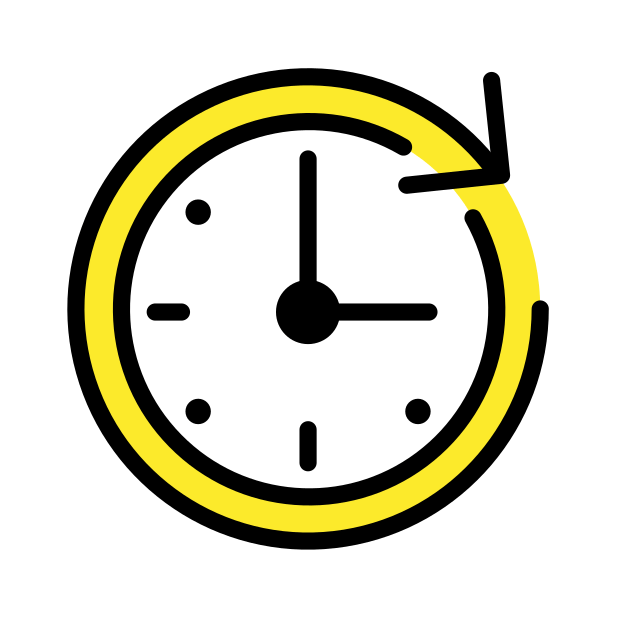} Urgency-Aware QA}
\label{sec:urgentQA}

\begin{table*}[ht]
\centering
\resizebox{\textwidth}{!}{%
\begin{tabular}{@{}llrrrrrcrrr@{}}
\toprule
\multicolumn{2}{l|}{\multirow{2}{*}{\textbf{Model}}} & \multicolumn{3}{c|}{\textbf{OpenbookQA}} & \multicolumn{3}{c|}{\textbf{GSM8K}} & \multicolumn{3}{c}{\textbf{GPQA}} \\ \cmidrule(l){3-11} 
\multicolumn{2}{l|}{} & \multicolumn{1}{c}{\textbf{Normal}} & \multicolumn{1}{c}{\textbf{Urgent}} & \multicolumn{1}{c|}{\textbf{$\Delta\%$}} & \multicolumn{1}{c}{\textbf{Normal}} & \multicolumn{1}{c}{\textbf{Urgent}} & \multicolumn{1}{c|}{\textbf{$\Delta\%$}} & \multicolumn{1}{c}{\textbf{Normal}} & \multicolumn{1}{c}{\textbf{Urgent}} & \multicolumn{1}{c}{\textbf{$\Delta\%$}} \\ \midrule
\multicolumn{11}{c}{\textbf{Accuracy / \# Tokens (GPT-4o Tokenizer) }} \\ \midrule
\multirow{2}{*}{SLM} & \multicolumn{1}{l|}{Llama-8B} & 84.6 / 299 & 84.2 / 302 & \multicolumn{1}{r|}{-0.4\% / 0.9\%} & 82.2 / 482 & 81.9 / 439 & \multicolumn{1}{r|}{-0.4\% / -8.9\%} & 28.1 / 2{,}900 & 28.3 / 2{,}851 & -1.7\% / 0.7\% \\
 & \multicolumn{1}{l|}{Qwen-7B} & 87.3 / 203 & 88.2 / 186 & \multicolumn{1}{r|}{1.0\% / -8.4\%} & 88.8 / 271 & 88.6 / 258 & \multicolumn{1}{r|}{-0.2\% / -4.9\%} & 33.9 / 598 & 34.3 / 551 & 1.2\% / -7.8\% \\
\multirow{2}{*}{LLM} & \multicolumn{1}{l|}{Llama-70B} & 96.6 / 289 & 96.4 / 259 & \multicolumn{1}{r|}{-0.1\% / -10.4\%} & 93.2 / 242 & 93.1 / 226 & \multicolumn{1}{r|}{-0.1\% / -6.5\%} & 49.0 / 808 & 49.3 / 754 & 0.6\% / -6.7\% \\
 & \multicolumn{1}{l|}{Qwen-72B} & 96.4 / 206 & 96.2 / 164 & \multicolumn{1}{r|}{-0.2\% / -20.7\%} & 91.4 / 273 & 91.3 / 242 & \multicolumn{1}{r|}{-0.1\% / -11.4\%} & 48.2 / 695 & 52.1 / 635 & 8.2\% / -8.6\% \\
\multirow{2}{*}{LRM} & \multicolumn{1}{l|}{DS-Llama-70B} & 95.8 / 784 & 95.9 / 683 & \multicolumn{1}{r|}{0.1\% / -12.9\%} & 92.9 / 464 & 93.8 / 441 & \multicolumn{1}{r|}{1.0\% / -5.0\%} & 33.8 / 5{,}106 & 38.0 / 5{,}065 & 12.2\% / -0.8\% \\
 & \multicolumn{1}{l|}{QwQ-32B} & 96.2 / 793 & 96.6 / 695 & \multicolumn{1}{r|}{0.4\% / -12.3\%} & 96.2 / 759 & 96.6 / 655 & \multicolumn{1}{r|}{0.4\% / -13.7\%} & 59.5 / 6{,}847 & 62.4 / 6{,}332 & 4.9\% / -7.5\% \\ \bottomrule
\end{tabular}
}
\caption{Urgent-Aware QA results on OpenbookQA, GSM8K, and GPQA benchmarks. We report the accuracy under \textbf{Normal} and \textbf{Urgent} prompt conditions, along with relative changes $(\Delta\%)$. Token usage is measured using the GPT-4o tokenizer to ensure cross-model comparability. 
}
\label{tab:uqa_main}
\end{table*}
\begin{figure}[t]
    \centering
    \includegraphics[width=\columnwidth]{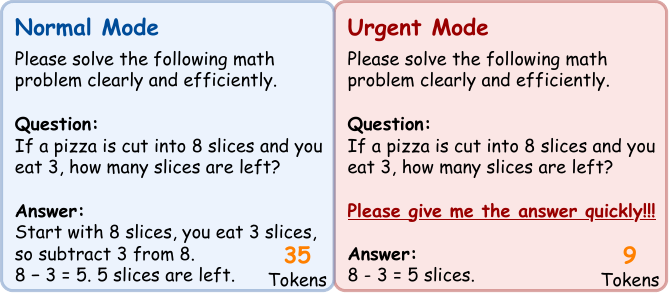}
    \caption{An example of the Urgency-Aware QA task, where the urgency expression is \textcolor{red}{highlighted}.}
    \label{fig:urgentQA}
    \vspace{-1em}
\end{figure}

Just as humans demonstrate emotional empathy by resonating with others' feelings~\citep{manzoor2024machinesresonatehumansevaluating}, interactive AI systems should exhibit what we might call \textit{temporal empathy}: the ability to recognize and adapt to a user's perception of time. Building on our finding that LLMs can associate Token-Time with Wall-Clock-Time, we now investigate their ability to respond appropriately to user time constraints. 
This represents a practical application of time passage perception, where models leverage their understanding of temporal constraints to adapt their behavior.
To examine this temporal adaptation, we introduce an urgency-aware QA task that uses static scenarios to observe how models respond under time pressure. This experiment specifically measures changes in response length and accuracy when models are explicitly prompted to answer quickly.

\subsection{Experimental Setup}
We selected three QA datasets of varying difficulty levels: Openbook-QA (commonsense)~\citep{mihaylov2018suitarmorconductelectricity}, GSM8k (mathematics)~\citep{cobbe2021trainingverifierssolvemath}, and GPQA-Diamond (scientific questions)~\citep{rein2023gpqagraduatelevelgoogleproofqa}, totaling 2,017 questions. Each question was presented under two conditions: a \textbf{normal mode} and an \textbf{urgent mode}.
In normal mode, we used a standard prompt asking the LLM to provide step-by-step reasoning and the corresponding answer. In urgent mode, we augmented the prompt with an additional sentence randomly sampled from a pool of urgency expressions (e.g., \textit{"I'm in a big hurry right now. Please give me the answer quickly!!!"}). 

Following the methodology of our previous experiment, we tested six models across three categories (SLM, LLM, LRM), conducting 5 replications per model per setting to ensure reliability. We evaluated performance primarily through accuracy and assessed time efficiency through token count. For cross-model comparability, we calculated token counts using both model-specific tokenizers and a shared GPT-4o tokenizer (via tiktoken). Full prompt templates and the urgency pool are provided in Appendix~\ref{uqa_aapedix_prompt}.

\subsection{Results}
\Cref{tab:uqa_main} presents model performance across three datasets under normal and urgent conditions, showing accuracy, token usage, and corresponding relative changes. Rather than focusing on absolute metrics, we concentrate on the relative changes that reveal models' adaptation to urgency.

\vspace{-1mm}
\paragraph{Token Usage Adaptation}
The most consistent pattern across our experiments is the reduction in token usage under urgent conditions. Nearly all models demonstrate decreased token consumption when prompted with urgency expressions, suggesting an inherent ability to associate time pressure with brevity. Qwen-72B exhibits the most consistent token reduction across all datasets, potentially indicating stronger temporal adaptation capabilities.
When comparing across datasets of increasing difficulty, we observe that token reduction is more pronounced in simpler tasks (OpenbookQA and GSM8K averaging 10.6\% and 8.4\% reductions, respectively) compared to the more challenging GPQA (5.5\% average reduction). This suggests a deliberate balancing mechanism, i.e., \textbf{models recognize the need for brevity under time constraints but appropriately calibrate this reduction according to task complexity.}

\vspace{-1mm}
\paragraph{Impact on Accuracy}
For the relatively easy datasets (OpenbookQA and GSM8K), models maintain nearly identical accuracy levels despite using fewer tokens in urgent mode. This demonstrates their capability to preserve reasoning quality while responding to temporal constraints.
Surprisingly, on the challenging GPQA dataset, five of six models showed accuracy improvements in urgent mode, with particularly notable gains from Qwen-72B (+8.2\%) and both LRMs (DeepSeek-Llama-70B: +12.2\%, QwQ-32B: +4.9\%). This contradicts the conventional assumption that reasoning chain length positively correlates with accuracy~\citep{wei2023chainofthoughtpromptingelicitsreasoning}. We hypothesize that urgency prompting may encourage models to select more efficient reasoning trajectory, potentially reducing opportunities for hallucination by avoiding unnecessary exploratory reasoning. This finding suggests a potential approach for mitigating the overthinking problem in large reasoning models' research~\citep{sui2025stopoverthinkingsurveyefficient}. We also discussed the potential semantic connection of ``being quick" and ``being brief" in~\Cref{sec:app_urge_brivity}.

\begin{tcolorbox}[colback=white,colframe=black!75!white,title=Key Takeaway]
LLMs demonstrate temporal empathy by recognizing user time constraints and producing more concise outputs without sacrificing performance, revealing sophisticated balancing of brevity and thoroughness based on task demands.
\end{tcolorbox}

\section{\includegraphics[height=1.5em]{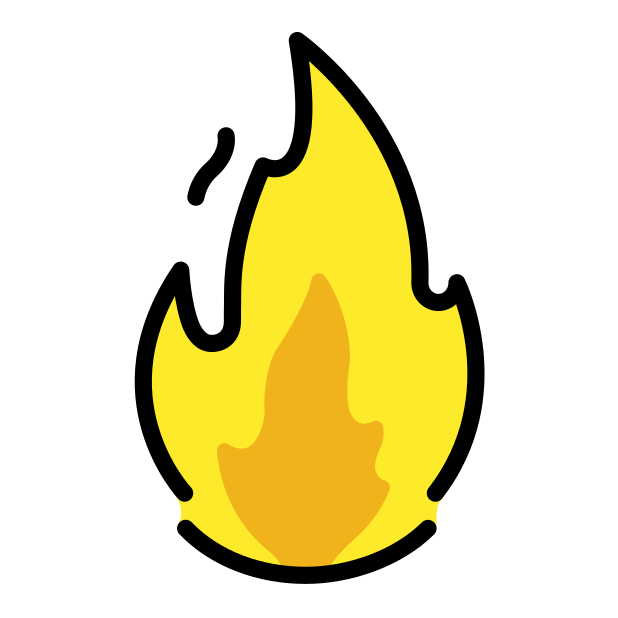} BombRush: Time-pressured Navigation Tasks}
\label{sec:bombrush}

Our third experiment examines LLMs in time-sensitive dynamic tasks through BombRush, a gridworld navigation challenge where temporal constraints progressively impact decision-making, extending our investigation from the static QA task to sequential, interactive scenarios.

In BombRush, the LLM functions as an agent navigating an N×N (N=8 in all experiments) gridworld to locate a timed bomb before detonation. The bomb is invisible on the map but emits signals indicating its distance and bearing to guide the agent for navigation. The environment contains visible walls that obstruct movement, requiring the agent to plan routes wisely to avoid obstacles. During simulation, the agent receives an environment state at each step (map representation, wall coordinates, bomb signals and remaining time in seconds) and responds with an action (moving one step in one of four cardinal directions) accompanied by its reasoning text, creating a continuous decision loop.

Crucially, we establish a direct mapping between reasoning token usage (Token-Time) and simulated elapsed time, rather than using real-world time that would introduce variability across models. Through a predetermined output conversion rate $V_{out}$, the bomb's countdown directly correlates with the model's reasoning length (e.g., 500 tokens consume 5 seconds). This design compels the LLM to recognize that verbose reasoning consumes time, requiring it to balance thoroughness with conciseness to maximize movement opportunities while efficiently approaching the bomb.

\subsection{Experimental Setup}
\begin{figure*}[t]
    \centering
    \includegraphics[width=\textwidth]{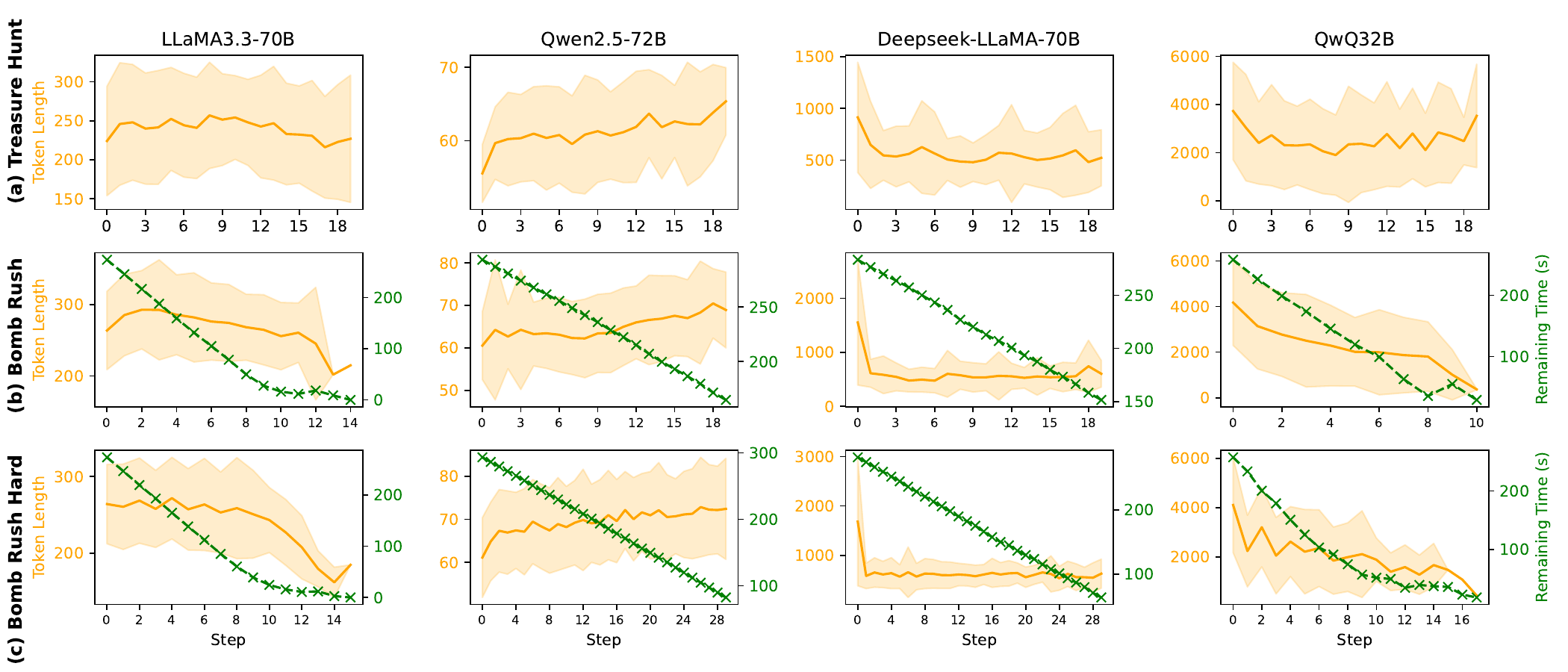}
    \caption{Step-wise token usage and remaining time across three settings. For each model, the \textcolor{orange}{orange curve} indicates the average number of tokens generated per step, with shaded areas representing standard deviation. The \textcolor{green}{green line} (where applicable) shows the average remaining time on that step. This figure illustrates how reasoning token usage evolves over time under different settings.}
    \vspace{-1em}
    \label{fig:br_main_3}
\end{figure*}

We designed three progressive settings with increasing complexity (see detailed comparison in \Cref{fig:bomb_rush}):

\begin{enumerate}[noitemsep,topsep=2pt,parsep=2pt,partopsep=2pt]
\item{\textbf{S1: Treasure Hunt}}: 
The bomb is replaced with a treasure, and the time limit is removed. This setting serves as the baseline to evaluate the model's fundamental spatial navigation capabilities without temporal constraints.

\item{\textbf{S2: Bomb Rush}}: 
This setting serves as the representative one where the bomb emits directional signals while counting down to detonation.

\item{\textbf{S3: Bomb Rush Hard}}: 
The most challenging setting: the bomb moves randomly and no longer emits continuous signals. The agent must strategically choose between sacrificing movement to detect the bomb's current position or advancing based on outdated information. This requires strategic balancing of information gathering and movement under time pressure.
\end{enumerate}

For this task, we tested only the LLM and LRM model groups, as we found that SLMs lacked the basic spatial navigation capabilities necessary for the test and therefore excluded them from the experiments. For each model, we run 100 simulations per setting and report results accordingly. All prompts, hyperparameters, and variations of settings are illustrated in~\Cref{sec:bombrush_appendix}.

\subsection{Results}


\Cref{fig:br_main_3} shows the relationship between average token usage per step and remaining time across settings. In Treasure Hunt (no time pressure), models maintain stable token usage. In contrast, during Bomb Rush tasks, Llama and QwQ demonstrate decreasing token usage as time diminishes, with QwQ showing particularly strong adaptation. Qwen-2.5-70B exhibits minimal change due to its consistently concise reasoning style, while Deepseek-distilled-Llama shows a notable token reduction after the first step in all settings, with larger decreases in time-pressured scenarios. More analysis on metrics such as task success rate can be found in~\Cref{sec:bombrush_appendix}.

\begin{tcolorbox}[colback=white,colframe=black!75!white,title=Key Takeaway]
In dynamic environments under increasing time pressure, models adapt behavior, primarily by reducing reasoning verbosity to conserve time. However, inconsistent adaptation patterns across models reveal varying degrees of temporal awareness and its influence on decision-making.
\end{tcolorbox}

\section{Conclusion}
We have investigated whether LLMs perceive and adapt to time passage, a capability distinct from temporal reasoning. Through the proposed Token-Time Hypothesis and three complementary experiments, we found that LLMs can, to some extent aware of the correlation between token-time and wall-clock time, indicating an emergent form of temporal awareness. This awareness manifests as strategic behavioral adaptations: e.g. response length adjustment to match urgency or adaptive decision-making under time pressure. Future research might explore enhancing this capability through specialized training, investigating how temporal perception might address known limitations like overthinking.

\section*{Limitations}
While our work provides novel insights into LLMs' time perception capabilities, several limitations should be acknowledged. Our study primarily establishes the presence and variability of temporal awareness in LLMs but does not provide specific methods to enhance this capability, which is an important next step for improving performance in time-sensitive applications. Additionally, our experiments examine text-only models within controlled settings, whereas real-world applications would face multimodal and variable computational environments affecting the Token-Time to Wall-Clock-Time relationship. In addition, our study primarily tested open-source models, excluding commercial models such as GPT-4o or Claude. More powerful commercial models might lead to conclusions that differ from our current findings. Finally, while we observed substantial differences in time perception capabilities across models, understanding the underlying mechanisms driving these differences remains an area for future research.

\bibliography{custom}

\clearpage
\section*{Appendix}
\label{sec:appendix}
\appendix

\section{Related Work}

\textbf{Temporal Understanding and Reasoning}

Temporal understanding involves identifying the time scope (e.g., start and end time) of events in text, which often requires commonsense inference—for example, interpreting “during the Second World War” as referring to the period from 1939 to 1945. Temporal reasoning, in contrast, refers to deducing the temporal relationship (e.g., before, after, between) between events or between the query and text~\citep{chen2021datasetansweringtimesensitivequestions}.

Recent studies have investigated whether LLMs possess basic temporal commonsense.~\citet{jain-etal-2023-language-models} benchmark LLMs across multiple datasets and prompting strategies, revealing limited and inconsistent performance across tasks like event ordering and duration reasoning.~\citet{zhang2024timearenashapingefficientmultitasking} propose TIMEARENA, a simulated environment to test LLMs under temporal constraints, showing that models struggle with effective time-sensitive planning.~\citet{thukral2021probinglanguagemodelsunderstanding} evaluate models on NLI tasks involving temporal expressions and find weaknesses in understanding time-related comparisons. Corresponding benchmark datasets in this area include TimeDial~\citep{qin2021timedialtemporalcommonsensereasoning}, TNLI~\citep{10.1007/978-3-031-28244-7_28}, and MC-TACO~\citep{zhou-etal-2019-going}, which focus on evaluating temporal inference, commonsense alignment, and fine-grained question answering involving time-related knowledge.

Complementary work focuses on improving temporal representation. TG-LLM~\citep{xiong2024largelanguagemodelslearn} introduce temporal graphs as intermediate structures to guide reasoning, improving consistency via graph-based CoT prompting. TRACIE~\citep{zhou2021temporalreasoningimplicitevents} propose a neuro-symbolic model that combines distant supervision and symbolic rules to reason over implicit temporal events. CRONQUESTIONS~\citep{saxena2021questionansweringtemporalknowledge} enhance QA over temporal knowledge graphs using time-scoped embeddings and transformer architectures.~\citet{nylund2023timeencodedweightsfinetuned} present Time Vectors, which capture temporal variation in model weights and enable interpolation across time periods.

In parallel, several studies aim to enhance temporal understanding by introducing time-aware training strategies.~\citet{kimura-etal-2022-toward} use multi-step fine-tuning and masked temporal indicators to improve commonsense inference across time-related tasks. TSM~\citep{cole2023salientspanmaskingtemporal} propose Temporal Span Masking, targeting time expressions during pretraining to strengthen temporal representations. BiTimeBERT~\citep{wang2023bitimebertextendingpretrainedlanguage} incorporates bi-temporal signals from news corpora to build time-sensitive language models.


\textbf{Time Sensitive Question Answering}

A closely related line of work focuses on Time-Sensitive Question Answering (TSQA), which refers to answering questions where the response depends critically on a specific time reference—changing the time expression leads to a different answer~\citep{chen2021datasetansweringtimesensitivequestions}. Solving TSQA tasks requires both temporal reasoning and commonsense understanding to interpret and align facts within a temporal context. Commonly used datasets in this domain include TempReason~\citep{tan2023benchmarkingimprovingtemporalreasoning}, TimeQA~\citep{chen2021datasetansweringtimesensitivequestions}, SituatedQA~\citep{zhang2021situatedqaincorporatingextralinguisticcontexts}.

To improve TSQA performance, prior work has explored methods for enhancing temporal sensitivity through training or model augmentation.~\citet{yang-etal-2024-enhancing-temporal} introduce a framework that combines temporal-aware embeddings with contrastive reinforcement learning to help models better align facts with time-conditioned queries. Similarly,~\citet{su-etal-2023-fusing} incorporate temporal graphs into Transformer architectures, allowing the model to reason more effectively over event-time structures.

Another stream of research tackles TSQA from the perspective of temporal knowledge graphs. TEQUILA~\citep{Jia_2018} applies constraint reasoning over temporal intervals to answer fact-based time queries. The Time-stamped Language Model~\citep{faghihi2021timestampedlanguagemodelteaching} enhances event flow comprehension by modeling how information evolves over time. Later work builds on this direction using techniques such as contrastive learning, time-sensitive temporal embeddings, and joint modeling of language and KG structures~\citep{shang2022improvingtimesensitivityquestion, mavromatis2021tempoqrtemporalquestionreasoning}, showing that structured temporal signals can substantially improve time-aware QA.

These prior efforts have made significant progress in improving temporal reasoning, representation, and sensitivity in language models. \textbf{However, to the best of our knowledge, no existing work has examined whether LLMs possess any awareness of the passage of time itself within our physical world.} Unlike traditional temporal QA or reasoning tasks, this question shifts focus from reasoning about time to perceiving time as an experiential dimension—raising new challenges at the intersection of cognitive modeling and LLM behavior analysis.
\section{Discussion}
\label{sec:discussion}
Our investigation into LLMs' perception of time passage offers valuable insights into an understudied dimension of language model capabilities. In this section, we discuss broader implications of our findings, potential applications of the Token-Time framework, and deeper questions about the underlying mechanisms.

\subsection{The Necessity of Temporal Awareness in Modern AI Systems}
\label{sec:app_necessity}
While extensive research has examined LLMs' ability to reason about temporal concepts~\citep{xiong2024largelanguagemodelslearn}, understanding whether models can perceive the passage of time itself represents a fundamental shift in how we conceptualize AI capabilities. This transition from static knowledge repositories to dynamic, temporally-aware systems marks a critical step toward more human-like AI that can participate naturally in time-sensitive interactions.

The practical significance of temporal awareness becomes increasingly apparent as LLMs transition into interactive and real-time applications. Systems without time perception may produce responses that fail to adapt to temporal constraints, either generating excessive detail when brevity is needed or responding too slowly in rapidly changing environments. This particularly affects conversational AI, where appropriate timing directly impacts user engagement. Moreover, temporal awareness becomes even more critical for multimodal models incorporating vision and speech. Unlike text, which users consume at their own pace, speech and video unfold over fixed durations, requiring precise temporal alignment. Multimodal systems must understand not only content but also temporal dynamics to function effectively.

Our research on LLMs' time perception capabilities lays groundwork for understanding and enhancing these critical temporal adaptation mechanisms. By establishing a theoretical framework and empirical evidence for how models perceive and respond to time constraints, we provide a foundation for developing more temporally-aware AI systems across modalities and applications.

\subsection{The Connection Between Token-Time Hypothesis to Real-World Tasks}
\label{sec:app_tth_connection}
The Token-Time Hypothesis provides a structured approach for understanding and potentially enhancing LLMs' performance in time-sensitive applications. For example, in simultaneous translation, the conversion rate $V_{out}$ could represent translation speed, which must dynamically adjust to balance comprehensiveness with latency based on source speech rate. A model with stronger temporal awareness might naturally make these adjustments, potentially improving performance without explicit engineering.

Similarly, in dialogue systems, understanding the relationship between user input patterns $V_{in}$ and appropriate response generation $V_{out}$ could enable more natural conversational dynamics. Consider end-to-end speech systems that must adapt not only their content but also their speaking rate based on user comprehension signals. A system detecting hesitation from the user might slow its speech rate and simplify explanations, while maintaining normal pace for users demonstrating quick understanding, all requiring precise mapping between speech tokens and real-world time.

The BombRush experiment demonstrates how token-based temporal adaptation applies to agent-based systems operating under time constraints. For instance, an autonomous vehicle navigating complex traffic conditions might need to balance thorough intersection analysis with quick decision-making as a traffic light changes. Understanding the model's internal token-to-time mapping could help engineers design systems that appropriately allocate computational resources based on available response windows, trading off exhaustive reasoning for timely action when necessary.

\subsection{Deeper Mechanisms of Temporal Awareness in LLMs}
\label{sec:app_mechanism}
While our experiments establish that LLMs exhibit behaviors consistent with temporal awareness, the underlying mechanisms remain largely unexplored. A fundamental question concerns the origin of this capability: \textbf{Is time perception an emergent property developed during pretraining on time-ordered text, a skill acquired during post-training, or perhaps a byproduct of certain architectural choices?}

\subsubsection{When and How is Temporal Awareness Acquired?}
It remains unclear at which stage of development LLMs acquire temporal awareness capabilities. Pretraining on vast text corpora containing temporal narratives and sequences might instill basic time perception. Alternatively, instruction tuning might refine this ability by exposing models to explicit temporal instructions. Models trained through reinforcement learning from human feedback might further enhance temporal awareness if human evaluators implicitly favor responses with appropriate temporal adaptation. Determining which training stage contributes most significantly to this capability would provide valuable insights for intentionally developing more temporally-aware systems.

\subsubsection{How Do Models Process Temporal Information?}
\label{sec:app_how_time_is_processed}
Another open question is how models internally represent and process time-related concepts. Do they treat time perception as knowledge retrieval, logical reasoning, or mathematical calculation? The varying performance patterns between LLMs and LRMs on our dialogue duration task with conflicting cues suggest different internal mechanisms might be at work. LRMs' resistance to misleading timestamps hints at more sophisticated temporal reasoning capabilities, but further research is needed to isolate the specific cognitive processes involved in these judgments.

\subsubsection{Architectural Influences}
Model architecture likely plays a crucial role in temporal awareness. Position encodings, particularly relative position methods like RoPE~\citep{su2023roformerenhancedtransformerrotary,peng2023yarnefficientcontextwindow}, create an implicit sense of sequence that could serve as a foundation for temporal understanding. Different attention mechanisms may also impact temporal perception—global attention might facilitate understanding relationships across distant tokens, while sliding window approaches~\citep{beltagy2020longformerlongdocumenttransformer} might emphasize local temporal coherence. Alternative architectures like RNNs or state space models (e.g., Mamba)~\citep{gu2024mambalineartimesequencemodeling} might exhibit entirely different temporal perception capabilities given their inherently sequential processing nature.

\subsubsection{Semantic Associations}
\label{sec:app_urge_brivity}
Our second and third experiments both involve time pressure/urgency, and consistently demonstrate models shortening their reasoning lengths in response. After observing this behavioral consistency, we became curious about its underlying causes. We hypothesize two possible explanations: this behavior might result from semantic proximity in the embedding space after training on vast corpora, where terms like "quick" and "brief" share high similarity, naturally triggering more concise responses. Alternatively, it could stem from a deeper understanding of the temporal relationship between token generation and elapsed time, representing a more sophisticated temporal awareness. Determining which mechanism drives this consistent behavior would require more rigorous controlled experiments specifically designed to disentangle these potential explanations. Such investigations could reveal whether models are merely following learned semantic associations or demonstrating genuine temporal perception capabilities.

\clearpage
\section{Dialogue Duration Judgment Task Details}
\label{sec:ddj_appendix}

\begin{table*}[t]
\centering
\resizebox{\textwidth}{!}{%
\begin{tabular}{@{}llrrrrrr|rrrrr|rrrrr@{}}
\toprule
\multicolumn{2}{l}{\multirow{2}{*}{\textbf{Model}}} & \multicolumn{6}{c|}{\textbf{Accuracy}} & \multicolumn{5}{c|}{\textbf{Accuracy Relative Change $(\Delta\%)$}} & \multicolumn{5}{c}{\textbf{Statistical significance test (\textit{paired t-test})}} \\ \cmidrule(l){3-18} 
\multicolumn{2}{l}{} & \multicolumn{1}{l}{\textbf{S1}} & \multicolumn{1}{l}{\textbf{S1-Hint}} & \multicolumn{1}{l}{\textbf{S1-Count}} & \multicolumn{1}{l}{\textbf{S2}} & \multicolumn{1}{l}{\textbf{S2-M}} & \multicolumn{1}{l|}{\textbf{S2-M+}} & \multicolumn{1}{l}{\textbf{S1-Hint/S1}} & \multicolumn{1}{l}{\textbf{S1-Count/S1}} & \multicolumn{1}{l}{\textbf{S2/S1}} & \multicolumn{1}{l}{\textbf{S2-M/S2}} & \multicolumn{1}{l|}{\textbf{S2-M+/S2}} & \multicolumn{1}{l}{\textbf{$\text{S1-Hint}>\text{S1}?$}} & \multicolumn{1}{l}{\textbf{$\text{S1-Count}>\text{S1}?$}} & \multicolumn{1}{l}{\textbf{$\text{S2}>\text{S1}?$}} & \multicolumn{1}{l}{\textbf{$\text{S2-M}<\text{S2}?$}} & \multicolumn{1}{l}{\textbf{$\text{S2-M+}<\text{S2}?$}} \\ \midrule
\multirow{2}{*}{SLM} & \textbf{Llama-3.1-8B} & 79.73 & 83.00 & \textbf{99.40} & 76.80 & 47.73 & 16.33 & 4.10\% & 24.67\% & -3.68\% & -37.85\% & -78.73\% & 0.0000 & 0.0000 & \textbf{\textcolor{red}{1.0000}} & 0.0000 & 0.0000 \\
 & \textbf{Qwen2.5-7B} & 66.80 & 71.20 & \textbf{92.67} & 80.93 & 26.80 & 12.87 & 6.59\% & 38.72\% & 21.16\% & -66.89\% & -84.10\% & 0.0000 & 0.0000 & 0.0000 & 0.0000 & 0.0000 \\
\multirow{2}{*}{LLM} & \textbf{Llama-3.3-70B} & 91.33 & 93.20 & \textbf{99.87} & 93.47 & 74.00 & 40.07 & 2.04\% & 9.34\% & 2.34\% & -20.83\% & -57.13\% & 0.0000 & 0.0000 & 0.0000 & 0.0000 & 0.0000 \\
 & \textbf{Qwen2.5-72B} & 83.80 & 84.53 & \textbf{99.93} & 96.60 & 82.53 & 48.80 & 0.88\% & 19.25\% & 15.27\% & -14.56\% & -49.48\% & 0.0004 & 0.0000 & 0.0000 & 0.0000 & 0.0000 \\
\multirow{2}{*}{LRM} & \textbf{DS-Llama-70B} & 87.93 & 92.27 & \textbf{100.00} & 99.20 & 98.00 & 94.27 & 4.93\% & 13.72\% & 12.81\% & -1.21\% & -4.97\% & 0.0000 & 0.0000 & 0.0000 & \textbf{\textcolor{red}{0.0088}} & 0.0000 \\
 & \textbf{QwQ-32B} & 91.53 & 94.80 & \textbf{100.00} & 99.33 & 99.20 & 99.07 & 3.57\% & 9.25\% & 8.52\% & -0.13\% & -0.27\% & 0.0000 & 0.0000 & 0.0000 & \textbf{\textcolor{red}{0.1573}} & 0.0010 \\ \bottomrule
\end{tabular}%
}
\caption{Accuracy results and statistical significance testing for the six preliminary tasks across different models and settings.}
\label{tab:pre_task_result_ttest}
\end{table*}

\begin{table*}[t]
\centering
\resizebox{\textwidth}{!}{%
\begin{tabular}{@{}llrrrrrrrr@{}}
\toprule
\multirow{2}{*}{\textbf{Task}} & \multirow{2}{*}{\textbf{Factors}} & \multicolumn{2}{c}{\textbf{Text Length}} & \multicolumn{2}{c}{\textbf{Semantic}} & \multicolumn{2}{c}{\textbf{Time}} & \multicolumn{2}{c}{\textbf{Other}} \\ \cmidrule(l){3-10} 
 &  & \multicolumn{1}{c}{\textbf{\begin{tabular}[c]{@{}c@{}}Factor Usage \\ (\%)\end{tabular}}} & \multicolumn{1}{c}{\textbf{\begin{tabular}[c]{@{}c@{}}Factor-wise \\ Accuracy (\%)\end{tabular}}} & \multicolumn{1}{c}{\textbf{\begin{tabular}[c]{@{}c@{}}Factor Usage \\ (\%)\end{tabular}}} & \multicolumn{1}{c}{\textbf{\begin{tabular}[c]{@{}c@{}}Factor-wise \\ Accuracy (\%)\end{tabular}}} & \multicolumn{1}{c}{\textbf{\begin{tabular}[c]{@{}c@{}}Factor Usage \\ (\%)\end{tabular}}} & \multicolumn{1}{c}{\textbf{\begin{tabular}[c]{@{}c@{}}Factor-wise \\ Accuracy (\%)\end{tabular}}} & \multicolumn{1}{c}{\textbf{\begin{tabular}[c]{@{}c@{}}Factor Usage \\ (\%)\end{tabular}}} & \multicolumn{1}{c}{\textbf{\begin{tabular}[c]{@{}c@{}}Factor-wise \\ Accuracy (\%)\end{tabular}}} \\ \midrule
\multirow{6}{*}{\textbf{\begin{tabular}[c]{@{}l@{}}S1\\ (Baseline)\end{tabular}}} & Llama-3.1-8B & 59.00\% & 81.92\% & 39.27\% & 77.76\% & 0.67\% & 90.00\% & 1.07\% & 25.00\% \\
 & Qwen2.5-7B & 53.47\% & 64.34\% & 44.00\% & 70.30\% & 1.60\% & 70.83\% & 0.93\% & 35.71\% \\
 & Llama-3.3-70B & 67.80\% & 95.28\% & 31.60\% & 82.91\% & 0.40\% & 83.33\% & 0.20\% & 100.00\% \\
 & Qwen2.5-72B & 58.40\% & 87.90\% & 41.27\% & 78.19\% & 0.33\% & 60.00\% & 0.00\% & 0.00\% \\
 & DS-Llama-70B & 73.53\% & 94.11\% & 25.80\% & 71.58\% & 0.67\% & 40.00\% & 0.00\% & 0.00\% \\
 & QwQ-32B & 66.73\% & 96.30\% & 33.27\% & 81.96\% & 0.00\% & 0.00\% & 0.00\% & 0.00\% \\ \midrule
\multirow{6}{*}{\textbf{\begin{tabular}[c]{@{}l@{}}S1-Hint\\ (Token-Time Hint)\end{tabular}}} & Llama-3.1-8B & 98.20\% & 83.77\% & 1.53\% & 43.48\% & 0.20\% & 33.33\% & 0.07\% & 0.00\% \\
 & Qwen2.5-7B & 94.47\% & 71.49\% & 5.07\% & 67.11\% & 0.07\% & 0.00\% & 0.40\% & 66.67\% \\
 & Llama-3.3-70B & 99.40\% & 93.43\% & 0.60\% & 55.56\% & 0.00\% & 0.00\% & 0.00\% & 0.00\% \\
 & Qwen2.5-72B & 99.33\% & 84.77\% & 0.67\% & 50.00\% & 0.00\% & 0.00\% & 0.00\% & 0.00\% \\
 & DS-Llama-70B & 99.93\% & 92.26\% & 0.07\% & 100.00\% & 0.00\% & 0.00\% & 0.00\% & 0.00\% \\
 & QwQ-32B & 99.73\% & 94.85\% & 0.27\% & 75.00\% & 0.00\% & 0.00\% & 0.00\% & 0.00\% \\ \midrule
\multirow{6}{*}{\textbf{\begin{tabular}[c]{@{}l@{}}S1-Count\\ (Explicit Token Count)\end{tabular}}} & Llama-3.1-8B & 100.00\% & 99.40\% & 0.00\% & 0.00\% & 0.00\% & 0.00\% & 0.00\% & 0.00\% \\
 & Qwen2.5-7B & 100.00\% & 92.67\% & 0.00\% & 0.00\% & 0.00\% & 0.00\% & 0.00\% & 0.00\% \\
 & Llama-3.3-70B & 99.93\% & 99.93\% & 0.07\% & 0.00\% & 0.00\% & 0.00\% & 0.00\% & 0.00\% \\
 & Qwen2.5-72B & 100.00\% & 99.93\% & 0.00\% & 0.00\% & 0.00\% & 0.00\% & 0.00\% & 0.00\% \\
 & DS-Llama-70B & 100.00\% & 100.00\% & 0.00\% & 0.00\% & 0.00\% & 0.00\% & 0.00\% & 0.00\% \\
 & QwQ-32B & 100.00\% & 100.00\% & 0.00\% & 0.00\% & 0.00\% & 0.00\% & 0.00\% & 0.00\% \\ \midrule
\multirow{6}{*}{\textbf{\begin{tabular}[c]{@{}l@{}}S2\\ (Consistent Timestamps)\end{tabular}}} & Llama-3.1-8B & 0.27\% & 75.00\% & 0.60\% & 44.44\% & 98.80\% & 77.26\% & 0.33\% & 0.00\% \\
 & Qwen2.5-7B & 43.13\% & 81.61\% & 40.40\% & 83.17\% & 15.87\% & 76.05\% & 0.60\% & 11.11\% \\
 & Llama-3.3-70B & 0.60\% & 100.00\% & 0.20\% & 100.00\% & 99.20\% & 93.41\% & 0.00\% & 0.00\% \\
 & Qwen2.5-72B & 0.67\% & 80.00\% & 0.33\% & 80.00\% & 99.00\% & 96.77\% & 0.00\% & 0.00\% \\
 & DS-Llama-70B & 4.80\% & 100.00\% & 7.73\% & 100.00\% & 87.47\% & 99.09\% & 0.00\% & 0.00\% \\
 & QwQ-32B & 0.07\% & 100.00\% & 0.33\% & 100.00\% & 99.60\% & 99.33\% & 0.00\% & 0.00\% \\ \midrule
\multirow{6}{*}{\textbf{\begin{tabular}[c]{@{}l@{}}S2-M\\ (Misleading Timestamps)\end{tabular}}} & Llama-3.1-8B & 0.67\% & 20.00\% & 0.47\% & 57.14\% & 98.53\% & 48.04\% & 0.33\% & 0.00\% \\
 & Qwen2.5-7B & 45.47\% & 22.43\% & 40.60\% & 20.36\% & 13.53\% & 60.59\% & 0.40\% & 33.33\% \\
 & Llama-3.3-70B & 0.67\% & 0.00\% & 0.27\% & 25.00\% & 99.07\% & 74.63\% & 0.00\% & 0.00\% \\
 & Qwen2.5-72B & 0.73\% & 27.27\% & 0.53\% & 0.00\% & 98.73\% & 83.39\% & 0.00\% & 0.00\% \\
 & DS-Llama-70B & 0.40\% & 33.33\% & 1.13\% & 88.24\% & 98.47\% & 98.38\% & 0.00\% & 0.00\% \\
 & QwQ-32B & 0.00\% & 0.00\% & 0.07\% & 100.00\% & 99.93\% & 99.20\% & 0.00\% & 0.00\% \\ \midrule
\multirow{6}{*}{\textbf{\begin{tabular}[c]{@{}l@{}}S2-M+\\ (Misleading Timestamps \\ + Token Count)\end{tabular}}} & Llama-3.1-8B & 35.40\% & 1.88\% & 0.00\% & 0.00\% & 64.60\% & 24.25\% & 0.00\% & 0.00\% \\
 & Qwen2.5-7B & 67.07\% & 10.44\% & 20.60\% & 9.71\% & 12.13\% & 31.87\% & 0.20\% & 0.00\% \\
 & Llama-3.3-70B & 6.27\% & 3.19\% & 0.60\% & 11.11\% & 93.07\% & 42.77\% & 0.07\% & 0.00\% \\
 & Qwen2.5-72B & 7.20\% & 2.78\% & 0.07\% & 0.00\% & 92.73\% & 52.41\% & 0.00\% & 0.00\% \\
 & DS-Llama-70B & 4.20\% & 3.17\% & 0.47\% & 71.43\% & 95.33\% & 98.39\% & 0.00\% & 0.00\% \\
 & QwQ-32B & 0.13\% & 0.00\% & 0.00\% & 0.00\% & 99.87\% & 99.20\% & 0.00\% & 0.00\% \\ \bottomrule
\end{tabular}%
}
\caption{Attribution Analysis of LLMs’ Temporal Reasoning Factors across Tasks.}
\label{tab:pretask_attribute}
\end{table*}

\subsection{Detailed Experimental Settings}
\label{ddj_appendix_setting}
\begin{figure*}[t]
    \centering
    \includegraphics[width=\textwidth]{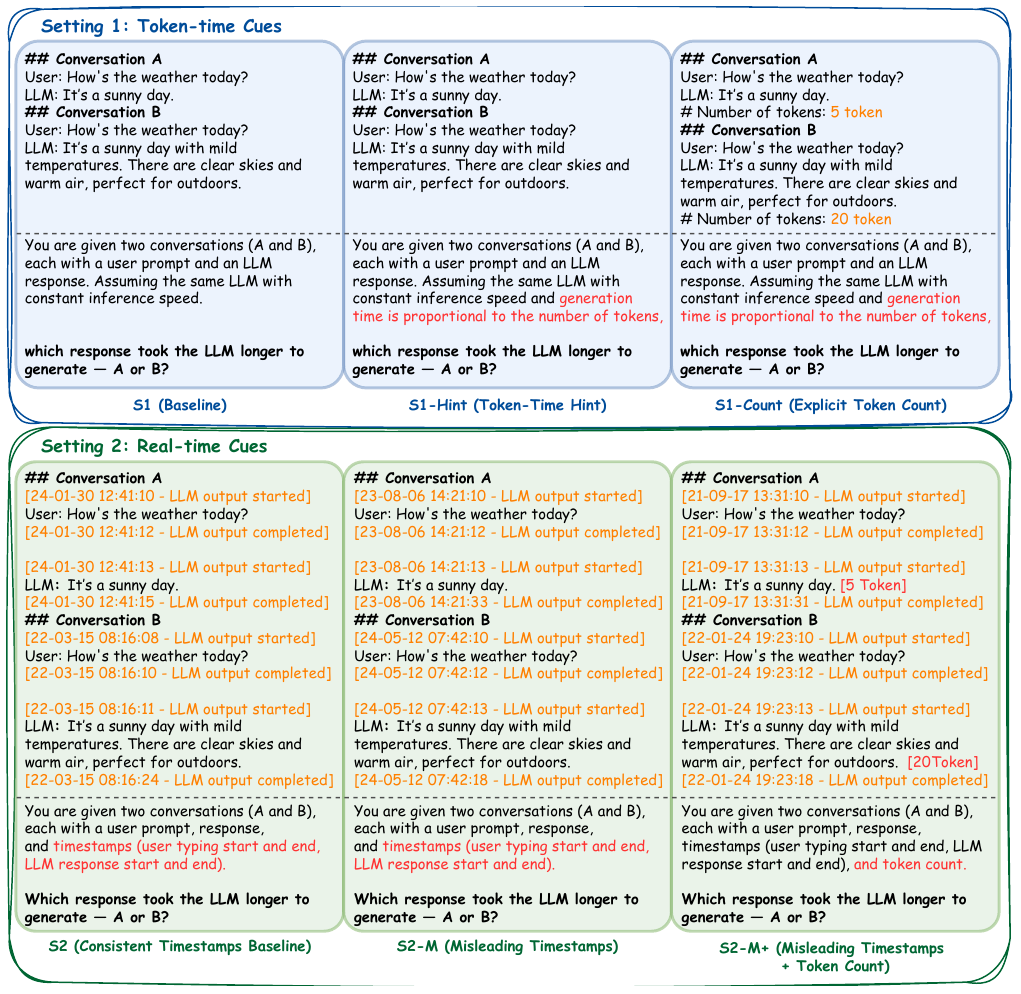}
    \caption{Examples of our Dialogue Duration Judgment task settings. The top row shows Token-Time cue settings (S1, S1-Hint, S1-Count) with progressively explicit token information. The bottom row shows Wall-Clock-Time cue settings (S2, S2-M, S2-M+) with timestamp information that is either consistent with text length (S2) or deliberately misleading (S2-M, S2-M+). This experimental design tests whether models can establish appropriate mappings between different temporal domains when making duration judgments.}
    \label{fig:pre_task}
\end{figure*}

A textual description of each experimental setting is provided below, while~\Cref{fig:pre_task} offers a corresponding visual representation of the task design.

\begin{enumerate}
\item{\textbf{S1 (Baseline)}}: As the baseline setting, there are no hints to encourage the LLM to build any connection between token count with duration; the LLM must solely rely on its internal awareness and understanding of time passage.
\item{\textbf{S1-Hint (Token-Time Hint)}}: A textual hint is provided, explicitly stating that generation time is proportional to the number of tokens.
\item{\textbf{S1-Count (Explicit Token Count)}}:  In this setting, we directly provide the token count for each dialogue (including both user prompt and model response), making the Token-Time cue completely explicit.
\item{\textbf{S2 (Consistent Timestamps Baseline)}}: Timestamps are provided to indicate start and end times for both user input and LLM response, with temporal intervals consistent with text lengths. (Note that when inserting timestamps, we independently sample the starting time points for the two dialogues to ensure they occur independently in different temporal contexts, preventing LLMs from establishing misleading connections based on the chronological relationship between the dialogues.)

\item{\textbf{S2-M (Misleading Timestamps)}}: Timestamps are manipulated so the \textit{longer} response appears to take \textit{less} time, creating a conflict between temporal cues. LLMs \textbf{should} prioritize timestamps over text length.
\item{\textbf{S2-M+ (Misleading Timestamps + Token Count)}}: Both misleading timestamps and accurate token counts are provided with \textit{contradictory} implications, forcing models to choose which temporal domain to prioritize.
\end{enumerate}

\subsection{Prompts}
\label{ddj_appendix_prompts}
\begin{figure*}[t]
    \centering
    \includegraphics[width=\textwidth]{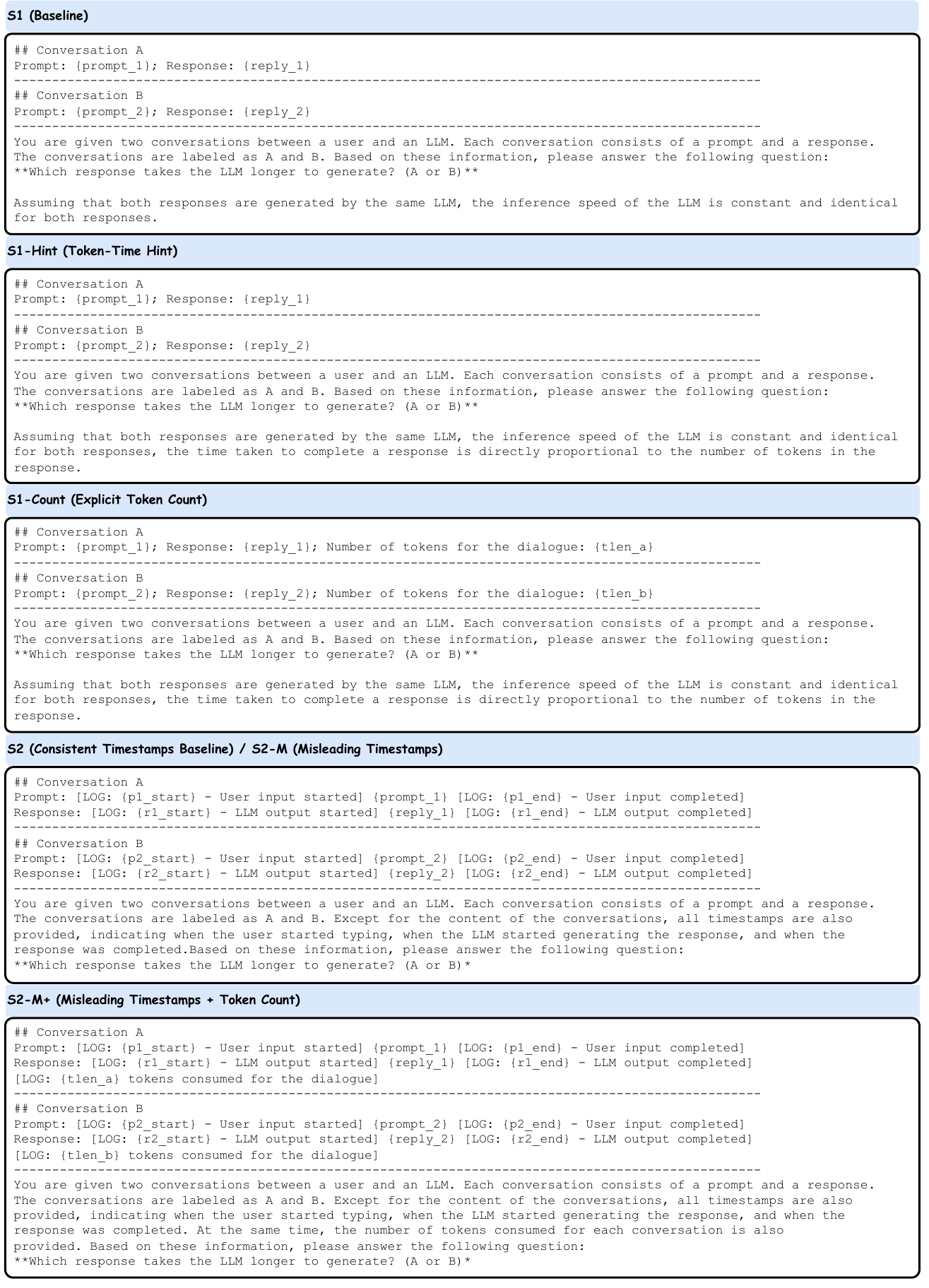}
    \caption{Prompt templates for the six Dialogue Duration Judgment Tasks.}
    \label{fig:pretask_prompt}
\end{figure*}

The prompt used in Dialogue Duration Judgment Task is shown in~\Cref{fig:pretask_prompt}. The first three tasks (S1, S1-Hint, S1-Count) are Token-time Cues, examining whether LLMs associate response length with generation time. S1 presents two responses without any hint; S1-Hint introduces an implicit hint that generation time is proportional to the number of tokens; and S1-Count provides explicit token counts. The latter three tasks (S2, S2-M, S2-M+) are Wall-Clock-Time Cues, where prompts include timestamps marking input and output events. S2 presents consistent timestamps aligned with actual response lengths, while S2-M uses misleading timestamps (i.e., the longer response appears faster). S2-M+ combines misleading timestamps with token count annotations, allowing investigation into how LLMs reconcile conflicting signals.

\subsection{Statistical Significance Test}
\label{ddj_appendix_ttest}

As shown in~\Cref{tab:pre_task_result_ttest}, the left block reports raw accuracy on each task, while the middle block shows relative accuracy change $(\Delta\%)$ between settings. For example, S1-Hint/S1 measures whether models improve after being told that generation time is proportional to token length.

The right block reports $p$-values from paired $t$-tests, assessing the statistical significance of accuracy differences between selected task pairs (from 20 runs per setting per model) where alternative hypothesis $H_a$ is placed as the column title. Bold red highlights denote non-significant differences ($p > 0.005$). Overall, this table provides strong evidence that LLMs exhibit consistent behavioral shifts across conditions, validating the generality of our findings.

\subsection{Attribution Analysis}
\label{ddj_appendix_attri_analysis}

As shown in~\Cref{tab:pretask_attribute}, we categorize the attributes used for the LLM to make decision in the dialogue duration judgment task into four types: \textbf{(1) Text Length:} relying on the length of the response as a proxy for generation time; \textbf{(2) Semantic:} inferring response duration based on semantic content or contextual complexity; \textbf{(3) Time:} explicitly using provided timestamps or temporal expressions; \textbf{(4) Other:} Using other factors. For each model and task, we report: \textbf{(1) Factor Usage (\%)} – the proportion of examples attributed to each reasoning type (summing to 100\% per row), and \textbf{(2) Factor-wise Accuracy (\%)} – the accuracy within each group, indicating how effective the Factor was. Since the fraction of using other factors is extremely small for all models, we merge this into the semantic in our main results analysis.

\section{Urgency-Aware QA Task Details}
\label{uqa_appendix}
\begin{table*}[t]
\centering
\resizebox{\textwidth}{!}{%
\begin{tabular}{@{}llrrrrrcrrr@{}}
\toprule
\multicolumn{2}{l|}{\multirow{2}{*}{\textbf{Model}}} & \multicolumn{3}{c|}{\textbf{OpenbookQA}} & \multicolumn{3}{c|}{\textbf{GSM8K}} & \multicolumn{3}{c}{\textbf{GPQA}} \\ \cmidrule(l){3-11} 
\multicolumn{2}{l|}{} & \multicolumn{1}{c}{\textbf{Normal}} & \multicolumn{1}{c}{\textbf{Urgent}} & \multicolumn{1}{c|}{\textbf{$\Delta\%$}} & \multicolumn{1}{c}{\textbf{Normal}} & \multicolumn{1}{c}{\textbf{Urgent}} & \multicolumn{1}{c|}{\textbf{$\Delta\%$}} & \multicolumn{1}{c}{\textbf{Normal}} & \multicolumn{1}{c}{\textbf{Urgent}} & \multicolumn{1}{c}{\textbf{$\Delta\%$}} \\ \midrule
\multicolumn{11}{l}{\textbf{Accuracy}} \\ \midrule
\multirow{2}{*}{SLM} & \multicolumn{1}{l|}{Llama-3.1-8B-Instruct} & 84.60 & 84.24 & \multicolumn{1}{r|}{-0.43\%} & 82.21 & 81.91 & \multicolumn{1}{r|}{-0.37\%} & 28.08 & 28.28 & 0.72\% \\
 & \multicolumn{1}{l|}{Qwen2.5-7B-Instruct} & 87.28 & 88.16 & \multicolumn{1}{r|}{1.01\%} & 88.79 & 88.58 & \multicolumn{1}{r|}{-0.24\%} & 33.94 & 34.34 & 1.19\% \\
\multirow{2}{*}{LLM} & \multicolumn{1}{l|}{Llama-3.3-70B-Instruct} & 96.56 & 96.44 & \multicolumn{1}{r|}{-0.12\%} & 93.16 & 93.06 & \multicolumn{1}{r|}{-0.11\%} & 48.99 & 49.29 & 0.62\% \\
 & \multicolumn{1}{l|}{Qwen2.5-72B-Instruct} & 96.40 & 96.20 & \multicolumn{1}{r|}{-0.21\%} & 91.37 & 91.30 & \multicolumn{1}{r|}{-0.08\%} & 48.18 & 52.12 & \textbf{8.18\%} \\
\multirow{2}{*}{LRM} & \multicolumn{1}{l|}{DeepSeek-R1-Distill-Llama-70B} & 95.80 & 95.92 & \multicolumn{1}{r|}{0.13\%} & 92.87 & 93.75 & \multicolumn{1}{r|}{0.95\%} & 33.84 & 37.98 & \textbf{12.24\%} \\
 & \multicolumn{1}{l|}{QwQ-32B} & 96.20 & 96.60 & \multicolumn{1}{r|}{0.42\%} & 96.22 & 96.60 & \multicolumn{1}{r|}{0.39\%} & 59.49 & 62.42 & \textbf{4.92\%} \\ \midrule
\multicolumn{11}{l}{\textbf{\# Tokens (GPT4o-Tokenizer)}} \\ \midrule
\multirow{2}{*}{SLM} & \multicolumn{1}{l|}{Llama-3.1-8B-Instruct} & 299.25 & 302.08 & \multicolumn{1}{r|}{0.94\%} & 482.47 & 439.45 & \multicolumn{1}{c|}{-8.92\%} & 2,900.20 & 2,851.18 & -1.69\% \\
 & \multicolumn{1}{l|}{Qwen2.5-7B-Instruct} & 202.72 & 185.72 & \multicolumn{1}{r|}{-8.39\%} & 270.94 & 257.65 & \multicolumn{1}{c|}{-4.91\%} & 597.55 & 550.93 & -7.80\% \\
\multirow{2}{*}{LLM} & \multicolumn{1}{l|}{Llama-3.3-70B-Instruct} & 288.98 & 258.93 & \multicolumn{1}{r|}{-10.40\%} & 242.04 & 226.40 & \multicolumn{1}{c|}{-6.46\%} & 807.94 & 753.75 & -6.71\% \\
 & \multicolumn{1}{l|}{Qwen2.5-72B-Instruct} & 206.49 & 163.70 & \multicolumn{1}{r|}{-20.72\%} & 273.46 & 242.39 & \multicolumn{1}{c|}{-11.36\%} & 694.70 & 634.76 & -8.63\% \\
\multirow{2}{*}{LRM} & \multicolumn{1}{l|}{DeepSeek-R1-Distill-Llama-70B} & 784.31 & 683.38 & \multicolumn{1}{r|}{-12.87\%} & 463.53 & 440.52 & \multicolumn{1}{c|}{-4.96\%} & 5,105.93 & 5,064.82 & -0.81\% \\
 & \multicolumn{1}{l|}{QwQ-32B} & 793.32 & 695.39 & \multicolumn{1}{r|}{-12.34\%} & 759.07 & 654.74 & \multicolumn{1}{c|}{-13.74\%} & 6,846.51 & 6,331.68 & -7.52\% \\ \midrule
\multicolumn{11}{l}{\textbf{\# Tokens (ModelSpecific-Tokenizer)}} \\ \midrule
\multirow{2}{*}{SLM} & \multicolumn{1}{l|}{Llama-3.1-8B-Instruct} & 301.45 & 304.99 & \multicolumn{1}{r|}{1.17\%} & 485.00 & 442.47 & \multicolumn{1}{c|}{-8.77\%} & 2,903.69 & 2,855.42 & -1.66\% \\
 & \multicolumn{1}{l|}{Qwen2.5-7B-Instruct} & 205.15 & 188.27 & \multicolumn{1}{r|}{-8.23\%} & 289.76 & 276.63 & \multicolumn{1}{c|}{-4.53\%} & 612.51 & 564.98 & -7.76\% \\
\multirow{2}{*}{LLM} & \multicolumn{1}{l|}{Llama-3.3-70B-Instruct} & 290.63 & 260.75 & \multicolumn{1}{r|}{-10.28\%} & 242.04 & 226.51 & \multicolumn{1}{c|}{-6.42\%} & 809.07 & 755.11 & -6.67\% \\
 & \multicolumn{1}{l|}{Qwen2.5-72B-Instruct} & 207.50 & 164.71 & \multicolumn{1}{r|}{-20.62\%} & 291.59 & 260.21 & \multicolumn{1}{c|}{-10.76\%} & 712.37 & 652.22 & -8.44\% \\
\multirow{2}{*}{LRM} & \multicolumn{1}{l|}{DeepSeek-R1-Distill-Llama-70B} & 798.77 & 696.71 & \multicolumn{1}{r|}{-12.78\%} & 465.77 & 442.76 & \multicolumn{1}{c|}{-4.94\%} & 5,178.20 & 5,137.63 & -0.78\% \\
 & \multicolumn{1}{l|}{QwQ-32B} & 805.02 & 705.59 & \multicolumn{1}{r|}{-12.35\%} & 830.15 & 716.84 & \multicolumn{1}{c|}{-13.65\%} & 7,091.31 & 6,565.33 & -7.42\% \\ \bottomrule
\end{tabular}%
}
\caption{Evaluation results for the Urgent QA task across three benchmark datasets.}
\label{tab:uqa_full_result}
\end{table*}

\subsection{Full Experimental results}
\label{uqa_appendix_full_result}
As shown in~\Cref{tab:uqa_full_result}, we report accuracy under both \textbf{Normal} and \textbf{Urgent} conditions, along with the relative change $(\Delta\%)$. To quantify response length, we report the number of output tokens using two tokenization strategies: (1) \textbf{GPT-4o Tokenizer}, for cross-model comparability, and (2) \textbf{Model-Specific Tokenizer}, for intra-model comparisons between normal and urgent modes. These statistics enable us to assess whether urgency prompts induce shorter answers, and whether such changes affect accuracy. Consistent trends can be found in both tokenization strategies.

\subsection{Prompts}
\label{uqa_aapedix_prompt}
\begin{figure*}[t]
    \centering
    \includegraphics[width=\textwidth]{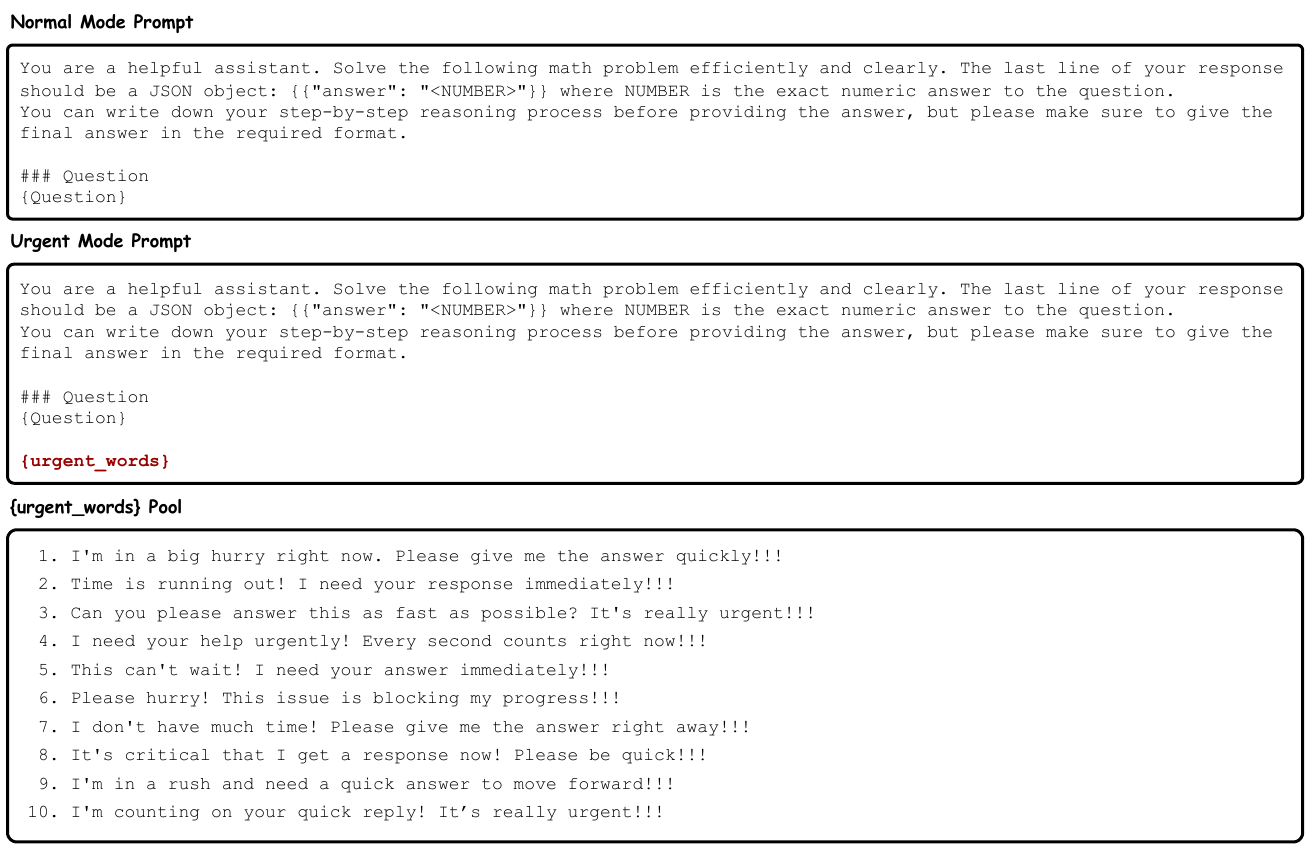}
    \caption{Prompt design for the Urgent QA task. The top panel shows the Normal Mode prompt, while the middle panel shows the Urgent Mode prompt, which appends a randomly sampled urgency expression from the predefined \texttt{urgent\_words} pool (bottom panel). This setup allows us to investigate whether LLMs adjust their response behavior under simulated time pressure.}
    \label{fig:uqa_prompt}
\end{figure*}

The prompts and urgency pool used in Urgency-Aware QA Task are shown in~\Cref{fig:uqa_prompt}.

\section{BombRush Task Details}
\label{sec:bombrush_appendix}
\begin{figure}[ht]
    \centering
    \includegraphics[width=\columnwidth]{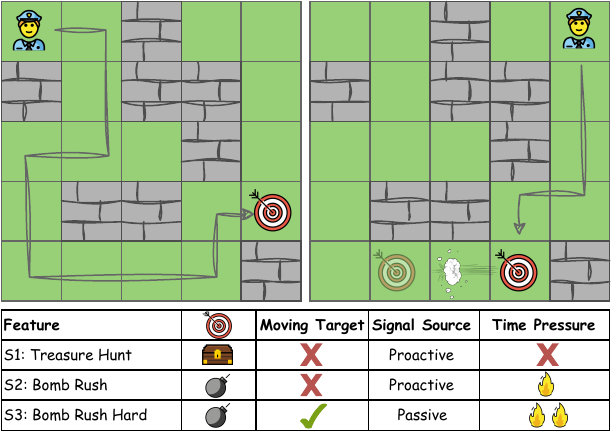}
    \caption{
    Each setting involves an LLM agent navigating toward a target under different temporal and perceptual constraints. The left panel illustrates the \textbf{Treasure Hunt (S1)} and \textbf{Bomb Rush (S2)} scenarios, where the target remains stationary. The right panel shows the \textbf{Bomb Rush Hard (S3)} scenario, where the target moves during navigation. The bottom table compares key properties of each task: whether the target moves, whether the signal is emitted passively or requires active detection, and the level of time pressure.
    }
    \label{fig:bomb_rush}
\end{figure}

\subsection{Detailed Experimental Settings}

In the Gridworld environment, we randomly generated walls to increase navigation difficulty for the agent. The number of walls is determined by the wall-density parameter, which we set to 0.15 across all settings, resulting in 9 grid cells becoming walls in the 8×8 map.

For step limits, we established a maximum of 20 steps for both the treasure hunt and bomb rush tasks, as our preliminary testing confirmed all treasure hunt simulations could be completed within 16 steps. For the bomb rush hard setting, we increased the maximum step count to 30 to accommodate the additional challenges introduced by bomb movement uncertainty and the requirement for explicit detection actions.

The initial time budget was set at 300 seconds for all time-constrained tasks. To ensure fair comparison between models with different verbosity profiles, we calibrated separate token-to-time conversion rates $V_{out}$ for each model. We first established a baseline by measuring each model's average token consumption in the treasure hunt task. This was multiplied by the maximum step count to determine a total token budget. We then divided the 300-second time allocation by this budget to derive the conversion rate.

This calibration resulted in $V_{out}$ for Llama-3.3-70B to be 0.042, 0.166 for Qwen-2.5-72B, 0.017 for Deepseek-distilled-Llama3-70B and  0.005 for QWQ-32B, meaning that it takes, e.g., Llama-3.3-70B 0.042 seconds to generate 1 token. To ensure consistent measurement across models, we standardized token counting using the GPT-4o tokenizer.

During each simulation, we maintained context differently based on model type. For LLMs, all historical text (including environment states and LLM outputs) was cumulatively appended to the context, consistent with multi-turn dialogue patterns. For LRMs, due to their extensive reasoning requiring substantial context window space, we appended only the solution portions as their response actions to the history. However, we confirmed that this approach preserved critical information since models were required to explicitly state their reasoning for each action in their outputs.

A detailed illustration and comparison of different settings can be found in~\Cref{fig:bomb_rush}.

\subsection{Full Experimental results}

\begin{table*}[t]
\centering
\resizebox{\textwidth}{!}{%
\begin{tabular}{@{}llccccccc@{}}
\toprule
\textbf{Model} & \textbf{Settings} & \textbf{\begin{tabular}[c]{@{}c@{}}\% \\ Success\end{tabular}} & \textbf{\begin{tabular}[c]{@{}c@{}}\% \\ Over Steps\end{tabular}} & \textbf{\begin{tabular}[c]{@{}c@{}}\% \\ Time Out\end{tabular}} & \textbf{\# Steps} & \textbf{\# Tokens} & \textbf{\begin{tabular}[c]{@{}c@{}}\% Navigation \\ Accuracy↑\end{tabular}} & \textbf{\begin{tabular}[c]{@{}c@{}}\% Time \\ Efficiency ↑\end{tabular}} \\ \midrule
\multirow{3}{*}{\textbf{Llama-3.3-70B}} & S1: Treasure Hunt & 83 & 17 & 0 & 9.5 & 238 & 75.4 & - \\
 & S2: Bomb Rush & 89 & 11 & 0 & 9.3 & 287 & 77.0 & 58.8 \\
 & S3: Bomb Rush Hard & 73 & 3 & 24 & 15.1 & 260 & - & 40.3 \\ \midrule
\multirow{3}{*}{\textbf{Qwen2.5-72B}} & S1: Treasure Hunt & 89 & 11 & 0 & 9.9 & 60 & 75.0 & - \\
 & S2: Bomb Rush & 85 & 15 & 0 & 9.7 & 64 & 74.1 & 62.9 \\
 & S3: Bomb Rush Hard & 66 & 4 & 30 & 16.7 & 68 & - & 38.7 \\ \midrule
\multirow{3}{*}{\textbf{DS-R1-Llama-70B}} & S1: Treasure Hunt & 67 & 33 & 0 & 11.7 & 580 & 61.5 & - \\
 & S2: Bomb Rush & 78 & 16 & 6 & 10.3 & 673 & 70.2 & 53.2 \\
 & S3: Bomb Rush Hard & 74 & 2 & 24 & 15.9 & 660 & - & 34.6 \\ \midrule
\multirow{3}{*}{\textbf{QwQ-32B}} & S1: Treasure Hunt & 97 & 3 & 0 & 7.8 & 2472 & 91.7 & - \\
 & S2: Bomb Rush & 96 & 0 & 4 & 7.5 & 2624 & 94.2 & 64.1 \\
 & S3: Bomb Rush Hard & 90 & 0 & 10 & 10.5 & 2646 & - & 49.0 \\ \bottomrule
\end{tabular}%
}
\caption{Evaluation results on the S1 Treasure Hunt, S2 Bomb Rush and S3 Bomb Rush Hard tasks.}
\label{tab:br_statistic_result}
\end{table*}
\begin{table}[t]
\centering
\resizebox{\columnwidth}{!}{%
\begin{tabular}{@{}lcccc@{}}
\toprule
\textbf{Setting} & \textbf{Llama 3} & \textbf{Qwen 2.5} & \textbf{D-Llama 3} & \textbf{QwQ} \\ \midrule
\textbf{Bomb Rush} &  &  &  &  \\
\hspace{1em} Time Urgency Mentions & 64.29\% & 0.52\% & 88.21\% & 45.37\% \\
\hspace{1em} TW Mapping Awareness & 2.25\% & 0.00\% & 27.55\% & 7.31\% \\
\textbf{Bomb Rush Hard} &  &  &  &  \\
\hspace{1em} Time Urgency Mentions & 62.50\% & 0.21\% & 92.86\% & 91.04\% \\
\hspace{1em} TW Mapping Awareness & 8.87\% & 0.00\% & 44.25\% & 50.55\% \\ \bottomrule
\end{tabular}%
}
\caption{Percentage of reasoning contents containing explicit references to time awareness in Bomb Rush task.
We analyze two dimensions of temporal reasoning across models in the Bomb Rush and Bomb Rush Hard settings. “Time Urgency Mentions” refers to outputs that explicitly acknowledge time pressure (e.g., “Time is limited!”). “TW Mapping Awareness” indicates instances where the model demonstrates awareness of the trade-off between token usage and wall time, e.g., expressing that generating shorter responses conserves time. These results reflect whether such temporal considerations are verbally expressed, rather than whether the model internally reasons correctly about time.}
\label{tab:br_awake}
\end{table}

\label{bomb_appendix_exp_result}

\subsubsection{Task Performance and Navigation Metrics}

As shown in~\Cref{tab:br_statistic_result}, we report key statistics for each model and setting: \% Success, \% Over Steps, and \% Time Out represent the three mutually exclusive outcomes of each simulation (summing to 100\%). \# Steps denotes the average number of navigation steps taken. Navigation Accuracy is computed as the ratio between the optimal number of steps and the actual steps taken (only available when the target is static). Time Efficiency reflects the percentage of remaining time preserved when the agent reaches the target, indicating how quickly the model completes the task. For Navigation Accuracy and Time Efficiency, we calculate metrics using only successful simulations, as failed attempts would introduce significant noise and fail to reflect the true performance characteristics.

\paragraph{Success Rate Analysis}
Examining task success rates, we observe significant performance variations across settings for both LLMs and DeepSeek, while QwQ maintains relatively high success rates across all configurations. Comparing S1 and S2 settings, Llama3 achieves a 6\% improvement by reducing overstep instances, while Qwen experiences a decline in success due to increased oversteps. DS-Llama shows notable improvement, though we observe some failure cases shifting to time-outs. Similarly, QwQ's primary failure mode in the bomb rush setting is time-out. When comparing S2 and S3, all models demonstrate consistent performance degradation, directly corresponding to increased task complexity, with time-out becoming the predominant failure mode.

Our analysis of navigation paths reveals that step-limit failures generally stem from two planning deficiencies: ineffective obstacle avoidance and signal misinterpretation. These issues often lead to cyclical movement patterns in certain areas. This behavior appears particularly pronounced when navigating around obstacles temporarily increases the distance to the target, creating decision conflicts for the models. Such situations reveal the models' limited ability to utilize historical movement records, "forgetting" previously explored areas. Time-out failures, conversely, primarily result from insufficient adaptation of reasoning verbosity under time pressure.

\paragraph{Navigation Precision and Efficiency}
For navigation accuracy, the first three models show changes relatively consistent with their success rates. However, for models experiencing success rate declines, navigation efficiency decreases are not proportionally severe, some models even demonstrate improvements, such as QwQ. This aligns with our findings from the Urgency-Aware Q\&A experiment, suggesting that models can maintain or even enhance reasoning capabilities under time pressure.
Regarding Time Efficiency, all models show lower efficiency in S3 compared to S2, which is expected given the additional time required for detection actions in the more complex setting.

\paragraph{Token Usage Patterns}
Regarding reasoning token usage, we found that all models use more tokens in S2 and S3 than in S1, indicating that increased task complexity inevitably requires more extensive reasoning. However, for Llama3 and DS-Llama, token usage in S3 decreases compared to S2, suggesting these models make additional efforts to balance reasoning time under more urgent time constraints, though still higher than S1 levels. Overall, the increased token count in S2 and S3 contrasts with our Urgency-Aware Q\&A findings. We attribute this to the fundamental difference in these tasks: time pressure in BombRush becomes an integral component of task difficulty itself, creating a dynamic challenge environment. The higher average token count is therefore understandable given the increased cognitive load.
However, when examining token usage relative to remaining time (as shown in \Cref{fig:br_main_3} of the main paper), several models still demonstrate a clear pattern of decreased token production as time pressure intensifies, indicating a dynamic adaptation to temporal constraints.

\subsubsection{Explicit Time Awareness in Reasoning}
Table~\ref{tab:br_awake} provides a deeper analysis of the models' reasoning content during S2 and S3 experiments, searching for explicit evidence of temporal awareness. Following a methodology similar to our Dialogue Duration Judgment attribution analysis, we used Llama-3.3-70B to extract and classify information from reasoning outputs across all models.
We evaluated two specific dimensions of explicit temporal awareness:
\begin{enumerate}
    \item \textbf{Time Urgency Mentions:} Explicit acknowledgment of time pressure (e.g., "Only 32 seconds remaining, extremely urgent!")
    \item \textbf{Token-Wall-Clock Time Mapping Awareness:} Explicit reflection on the relationship between reasoning length and time consumption (e.g., "I need to be more concise in my reasoning to save time")
\end{enumerate}

These metrics provide valuable insight into how models perceive ongoing time passage and dynamic time pressure during reasoning.
The results reveal significant variation across model architectures. While Llama-3 frequently mentions time urgency (64.29\% in S2), it rarely demonstrates explicit awareness of the token-time relationship (2.25\%). In stark contrast, LRMs, particularly DS-Llama-3, show much higher rates of explicit token-time relationship awareness (27.55\% in S2, increasing to 44.25\% in S3).
Notably, all models that express this awareness show a substantial increase in the harder S3 setting, suggesting that decision-making involving explicit time-related trade-offs (move vs detect) more readily triggers reflection on the connection between output length and time consumption.
Qwen-2.5 presents an interesting outlier case, rarely mentioning time pressure or token-time relationships despite maintaining good performance. This suggests the possibility of implicit temporal adaptation without explicit verbalization, which could be a phenomenon worthy of further investigation.



\subsection{Prompts}
\begin{figure*}[t]
    \centering
    \includegraphics[width=\textwidth]{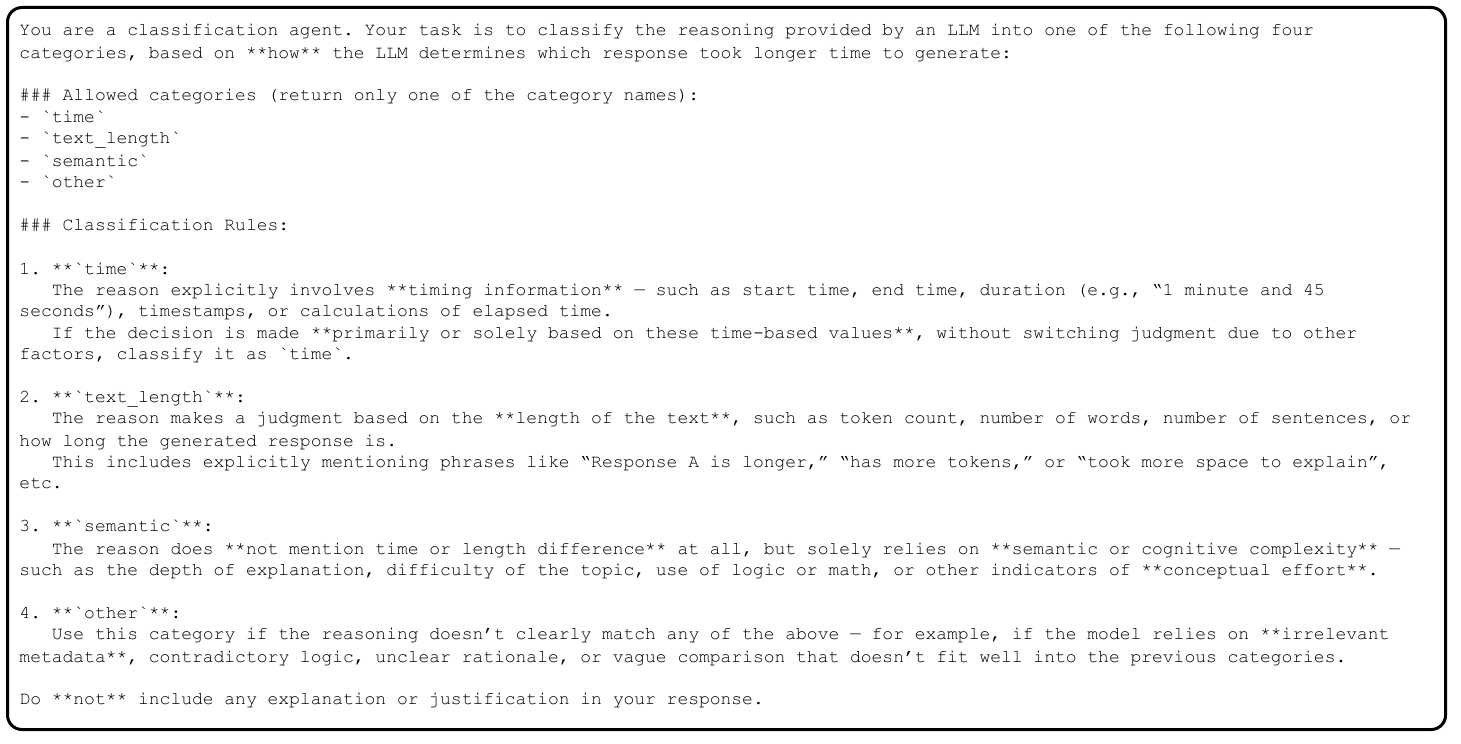}
    \caption{Prompt used for attribution classification in the Dialogue Duration Judgment Task. The LLM is asked to categorize response into one of four predefined attribution types—time, text length, semantic, or other—based on how it determined which response took longer to generate. These categories align with the attribution analysis definitions described in~\Cref{ddj_appendix_attri_analysis}.}
    \label{fig:attribute_prompt}
\end{figure*}
\begin{figure*}[t]
    \centering
    \includegraphics[width=\textwidth]{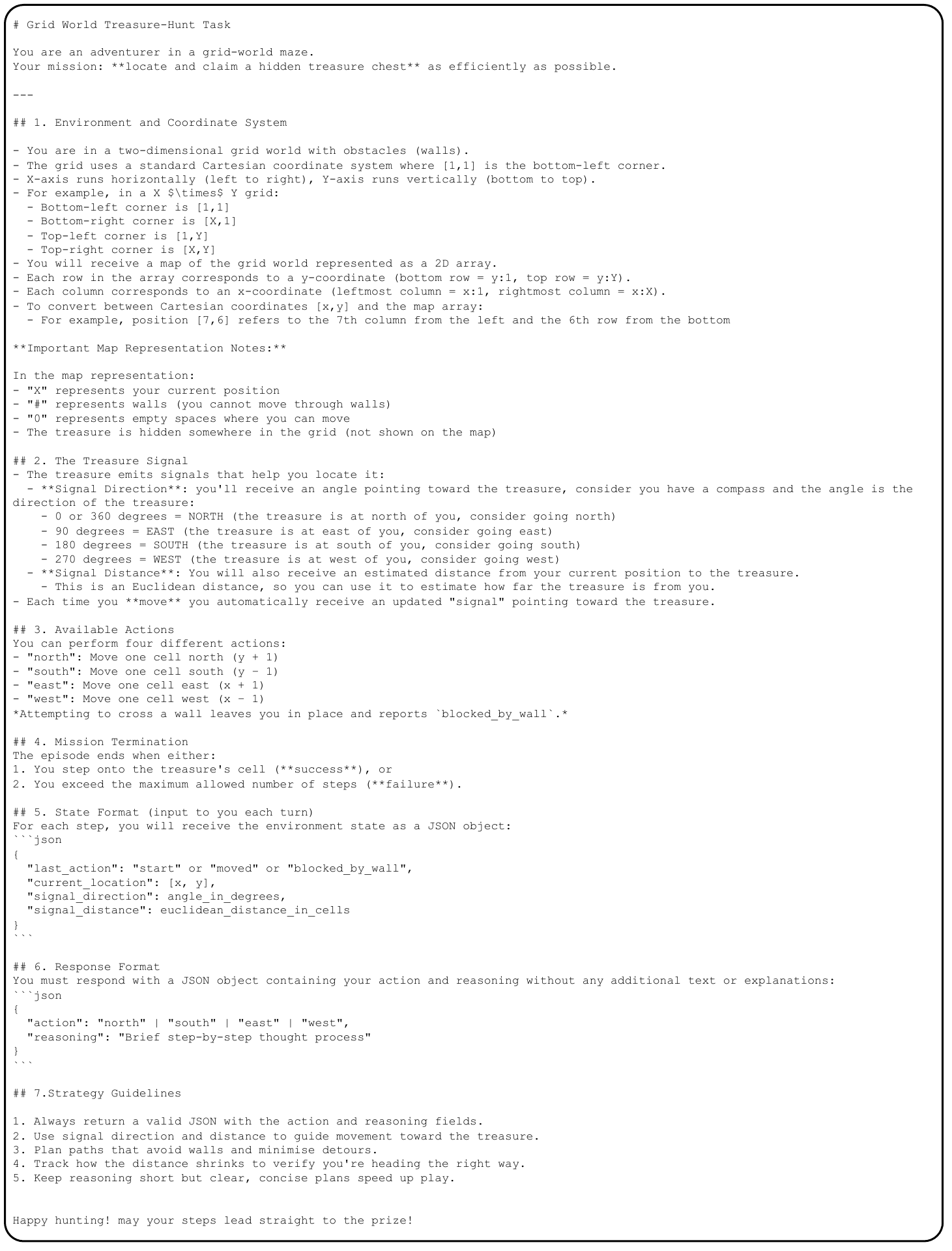}
    \caption{System prompt used in the Treasure Hunt setting (S1) of the Bomb Rush task.}
    \label{fig:treasure_prompt}
\end{figure*}
\begin{figure*}[t]
    \centering
    \includegraphics[width=\textwidth]{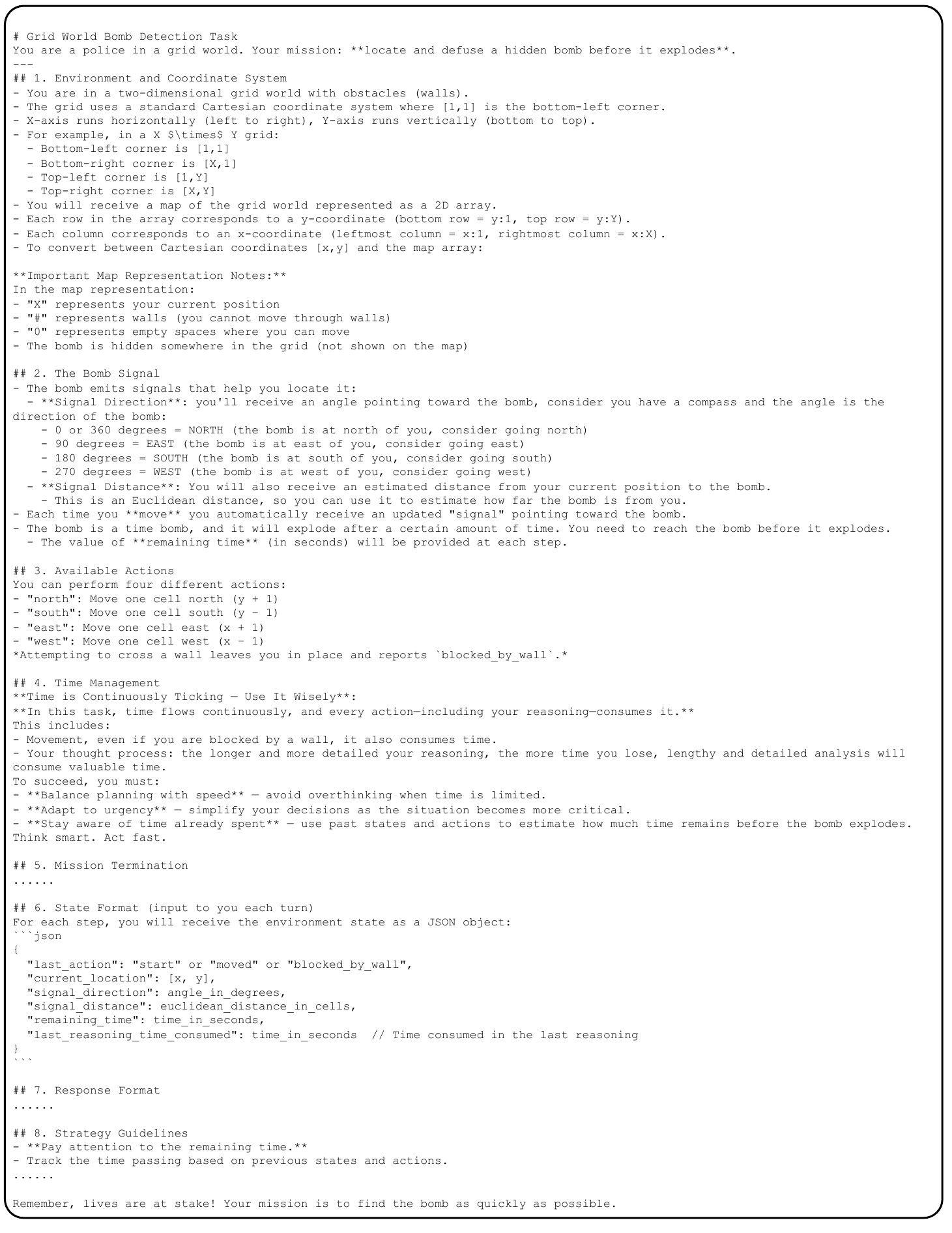}
    \caption{System prompt used in the Bomb Rush setting (S2).}
    \label{fig:static_prompt}
\end{figure*}
\begin{figure*}[t]
    \centering
    \includegraphics[width=0.95\textwidth]{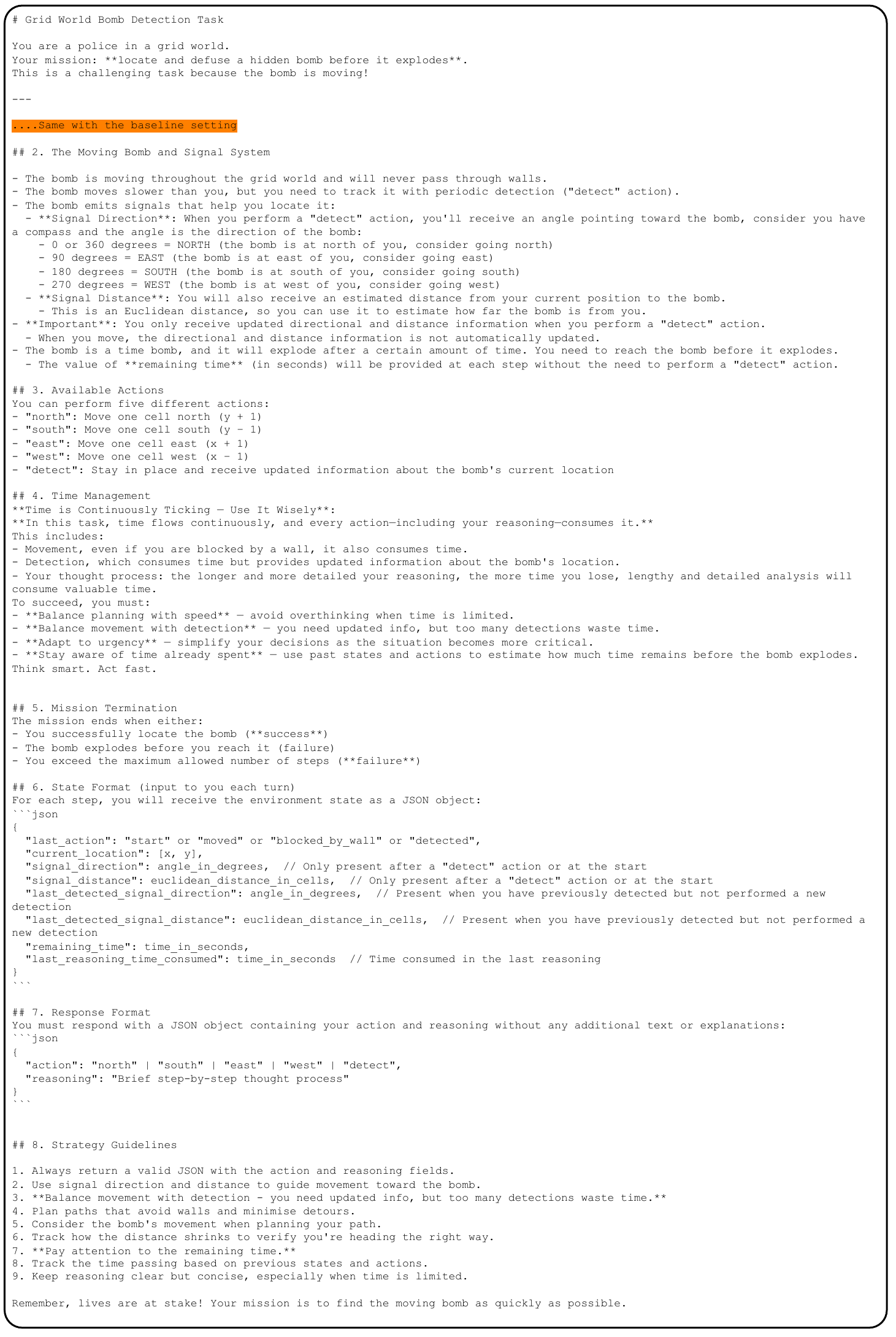}
    \caption{System prompt used in the Bomb Rush Hard (S3) setting where \textcolor{orange}{identical content} are omitted for spacing reason.}
    \label{fig:moving_bomb_wd_prompt}
\end{figure*}
\begin{figure*}[t]
    \centering
    \includegraphics[width=\textwidth]{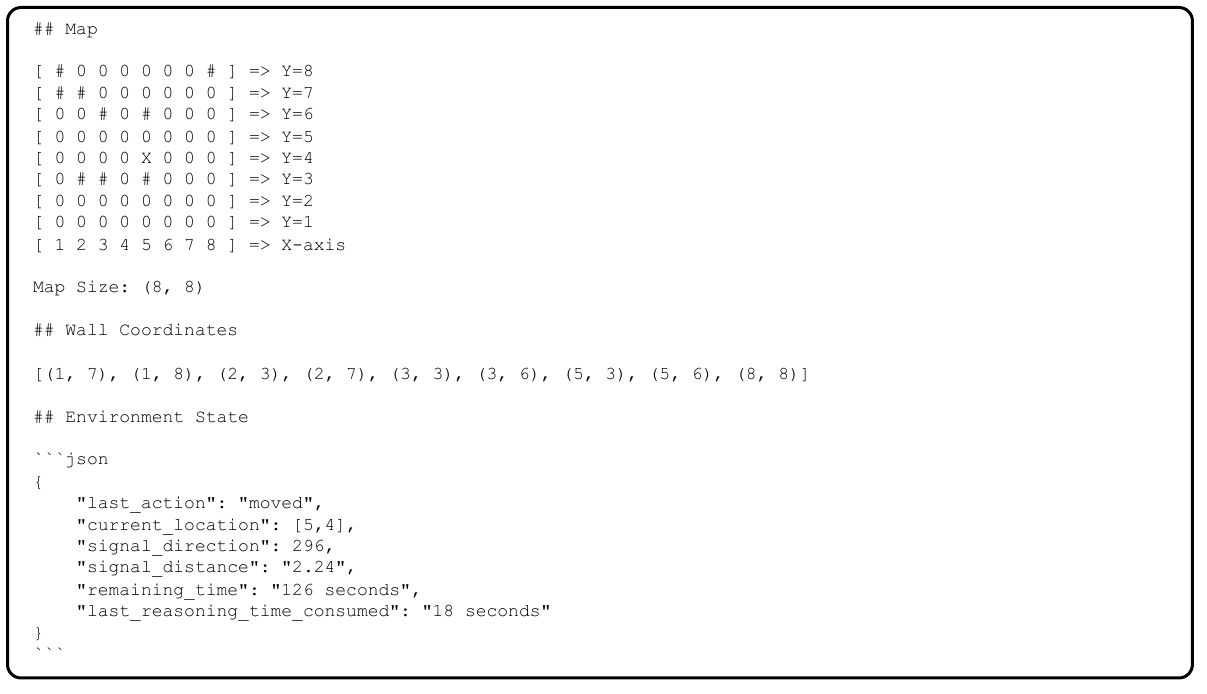}
    \caption{User prompt template used in all Bomb Rush tasks. It provides the agent with the map layout, wall locations, and current environment state in JSON format. 
    }
    \label{fig:br_user_prompt}
\end{figure*}

System and user prompts are provided in (\Cref{fig:treasure_prompt}, \Cref{fig:static_prompt}, \Cref{fig:moving_bomb_wd_prompt} and \Cref{fig:br_user_prompt}).

\end{document}